\documentclass[pmlr]{jmlr}% new name PMLR (Proceedings of Machine Learning)

\usepackage{booktabs}
\usepackage[load-configurations=version-1]{siunitx} % newer version
\usepackage[english]{babel} % for bibtex
\usepackage{subcaption}
\usepackage{soul}
\usepackage{makecell}
\usepackage{xcolor,colortbl}
\usepackage{comment}
\usepackage{longtable}% for long tables
\usepackage{float}

\theorembodyfont{\upshape}
\theoremheaderfont{\scshape}
\theorempostheader{:}
\theoremsep{\newline}

\jmlrvolume{106}
\jmlryear{2019}
\jmlrworkshop{Machine Learning for Healthcare}

\firstpageno{1}

\title[Effectively Modeling Time-Varying Relationships using RNNs]{Relaxed Parameter Sharing: Effectively Modeling Time-Varying Relationships in Clinical Time-Series}

\author{\Name{Jeeheh Oh}\footnotemark[1]
\Email{\href{mailto:jeeheh@umich.edu}{\color{black}{jeeheh@umich.edu}}}
\addr {\\ \footnotesize Department of Electrical Engineering and Computer Science \\
University of Michigan, Ann Arbor, MI, USA} 
\AND
\Name{Jiaxuan Wang}\footnotemark[1]
\Email{\href{mailto:jiaxuan@umich.edu}{\color{black}{jiaxuan@umich.edu}}} 
\addr {\\ \footnotesize Department of Electrical Engineering and Computer Science \\
University of Michigan, Ann Arbor, MI, USA} 
\AND
\Name{Shengpu Tang}
\Email{\href{mailto:tangsp@umich.edu}{\color{black}{tangsp@umich.edu}}}
\addr {\\ \footnotesize Department of Electrical Engineering and Computer Science \\
University of Michigan, Ann Arbor, MI, USA} 
\AND
\Name{Michael W. Sjoding}
\Email{\href{mailto:msjoding@med.umich.edu}{\color{black}{msjoding@med.umich.edu}}}
\addr {\\ \footnotesize Institute for Healthcare Policy \& Innovation; and \\
Department of Internal Medicine \\
University of Michigan, Ann Arbor, MI, USA}
\AND
\Name{Jenna Wiens}
\Email{\href{mailto:wiensj@umich.edu}{\color{black}{wiensj@umich.edu}}}
\addr {\\ \footnotesize Department of Electrical Engineering and Computer Science \\
University of Michigan, Ann Arbor, MI, USA} 
\AND
\footnotemark[1] \addr {\normalfont \footnotesize authors of equal contribution}
}

\begin{document}

\maketitle
% \vspace{-20pt}
\begin{abstract}%
Recurrent neural networks (RNNs) are commonly applied to clinical time-series data with the goal of learning patient risk stratification models. Their effectiveness is due, in part, to their use of parameter sharing over time (i.e., cells are repeated hence the name \textit{recurrent}). We hypothesize, however, that this trait also contributes to the increased difficulty such models have with learning relationships that change over time. Conditional shift, i.e., changes in the relationship between the input X and the output y, arises when risk factors associated with the event of interest change over the course of a patient admission. While in theory, RNNs and gated RNNs (e.g., LSTMs) in particular should be capable of learning time-varying relationships, when training data are limited, such models often fail to accurately capture these dynamics. We illustrate the advantages and disadvantages of complete parameter sharing (RNNs) by comparing an LSTM with shared parameters to a sequential architecture with time-varying parameters on prediction tasks involving three clinically-relevant outcomes: acute respiratory failure (ARF), shock, and in-hospital mortality. In experiments using synthetic data, we demonstrate how parameter sharing in LSTMs leads to worse performance in the presence of conditional shift. To improve upon the dichotomy between complete parameter sharing and no parameter sharing, we propose a novel RNN formulation based on a mixture model in which we relax parameter sharing over time. The proposed method outperforms standard LSTMs and other state-of-the-art baselines across all tasks. In settings with limited data, relaxed parameter sharing can lead to improved patient risk stratification performance. 
\end{abstract}

\section{Introduction}

Recurrent neural networks (RNNs) capture temporal dependencies between past inputs $\boldsymbol{x}_{1:t-1}$ and output $y_t$, in addition to the relationship between current input $\boldsymbol{x}_t$ and $y_t$. Their successful application to date is due in part to their explicit parameter sharing over time \citep{harutyunyan2017multitask, rajkomar2018scalable}. However, while advantageous in many settings, such parameter sharing could hinder the ability of the model to accurately capture time-varying relationships, i.e., tasks that exhibit temporal conditional shift. 

In healthcare, temporal condition shift may arise in clinical prediction tasks when the factors that put a patient at risk for a particular adverse outcome at the beginning of a hospital visit differ from those that put a patient at risk at the end of their stay. Failure to recognize conditional shift when building risk stratification models could lead to temporal biases in learned models; models may capture the average trend at the cost of decreased performance at specific points in time. This could be especially detrimental to models deployed and evaluated in real time. 

More formally, conditional shift refers to the change in the conditional distribution $P(Y=y|X=\boldsymbol{x})$ across tasks. In particular, we consider \textit{temporal} conditional shift, i.e., the setting in which the relationship between $\boldsymbol{x}$ and $y$ is a function of both $\boldsymbol{x}$ and time ($y_t=f(\boldsymbol{x},t;\theta_t)$). We hypothesize that RNN's complete sharing of parameters across time steps makes it difficult to accurately model temporal conditional shift. To address this, one could jointly learn a different cell for each time step, but such an architecture may easily lead to overfitting. More importantly, such an approach does not leverage the fact that many relationships are shared, at least in part, across time.

On synthetic data, in which we can control the amount of conditional shift, we explore the trade-offs in performance between models that share parameters across time versus models that do not. Beyond synthetic data, we illustrate the presence of temporal conditional shift in real clinical prediction tasks. To tackle this issue, we propose a novel RNN framework based on a mixture approach that relaxes parameter sharing over time, without sacrificing generalization performance. Applied to three clinically relevant patient risk stratification tasks, our proposed approach leads to significantly better performance relative to a long short-term memory network (LSTM). Moreover, the proposed approach can help shed light on task relatedness across time. 

\paragraph{Technical Significance.}
Our technical contributions can be summarized as follows:
\begin{itemize}  \setlength{\itemsep}{-5pt}
\item we formalize the problem setting of temporal conditional shift, 
\item we illustrate the presence of temporal conditional shift in three clinically relevant tasks
\item we propose a novel approach for relaxed parameter sharing within an RNN framework, and
\item we explore situations in which relaxed parameter sharing can help.
\end{itemize}
In theory, given enough data, RNNs should be able to accurately model relationships governed by temporal conditional shift. However, oftentimes in clinical applications, we have a limited amount of data to learn from. Going forward, researchers should check for the presence of conditional shift by comparing the proposed approach with an LSTM. If conditional shift is detected, then one may be able to more accurately model temporal dynamics through relaxed parameter sharing.

\paragraph{Clinical Relevance.}
Though tasks involving time-varying relationships are common in healthcare, current techniques rarely explicitly model temporal conditional shift.  In this work, we investigate the extent to which temporal conditional shift impacts clinical prediction tasks. We consider clinical tasks involving prediction of three adverse outcomes: acute respiratory failure (ARF), shock, and in-hospital mortality during an ICU admission. These tasks were selected based on their clinical relevance. ARF contributes to over 380,000 deaths in the US per year \citep{stefan2013epidemiology} and represents a challenging prediction task due to its multi-factorial etiology. Shock refers to the inadequate perfusion of blood oxygen to organs or tissues and can result in severe organ dysfunction and death when not recognized and treated immediately \citep{gaieski2016shock}. Both of these conditions are upstream events that contribute to patient risk of in-hospital mortality, our third prediction task. The ability to identify patients at risk of developing ARF or shock could facilitate improved patient triage and timely intervention, preventing irreversible damage and ultimately better patient outcomes. Finally, though we consider only these three tasks, we hypothesize that time-varying risk factors may arise in other clinical prediction tasks.

\section{Background \& Related Work} 

We focus on developing techniques that can handle temporal conditional shift, a type of data shift that commonly occurs in tasks involving clinical time-series data. There are two main types of data shift: i) covariate shift and ii) conditional shift. Covariate shift is the scenario where $P(X=\boldsymbol{x})$ varies across datasets \citep{reddi2015doubly, sugiyama2007covariate}, e.g., the distributions of patient demographics may differ across study populations. In contrast, conditional shift, our main focus, occurs when $P(Y=y|X=\boldsymbol{x})$ changes \citep{zhang2013domain, gong2016domain}, e.g., two hospitals may have similar patient populations, but different factors could drive patient risk due to differences in clinical protocols. In conditional shift, the relationship between input $\boldsymbol{x}$ and output $y$ has shifted. This can occur independently of a change in population. For some time, the study of data shift has driven research in the fields of domain adaptation, transfer learning, and multitask learning \citep{daume2009frustratingly, pan2010survey, ding2017recurrent, thiagarajan2018can}.   

Methods for dealing with conditional shift are largely driven by the problem setting. Researchers have explored the use of pre-trained features \citep{sharif2014cnn}, generalizable representations \citep{glorot2011domain, zhuang2015supervised}, and applying importance re-weighting techniques \citep{zhang2013domain}. In contrast to these works, we focus on techniques for tackling conditional shift in which the shift is driven by changes in time. In this setting, there is no clear distinction between tasks, because the change occurs gradually. Though related, this differs from `data drift' (i.e., the setting in which relationships change longitudinally) since we consider time on a local/relative scale as opposed to a global/absolute scale \citep{dos2016fast, soemers2018adapting}. That is, instead of focusing on differences between 2018 and 2019, we focus on changes within an admission or a patient. Though such local shift is expected to occur \citep{bellera2010variables, dekker2008survival}, it is often overlooked when modeling patient risk.

In the linear setting, past work has explored the use of multitask learning to model the temporal evolution of risk factors within a patient admission, where each day corresponds to a different model, but models are learned jointly \citep{wiens2016patient}. Related, \cite{dekker2008survival} proposed a variation to Cox regression analysis, studying different time windows separately and using time specific hazard ratios. 

Nonlinear methods designed to explicitly deal with temporal conditional shift have more recently been explored, focusing primarily on modifications to RNN architectures \citep{ha2016hypernetworks, park2017early}, especially LSTMs. For example, \cite{ha2016hypernetworks} proposed an extension to LSTMs, Hypernetwork, that relaxes parameter sharing by learning an auxiliary network that sets the primary network's parameters at each time step. Specifically, the auxiliary network can change the primary network's parameters through a scalar multiplier. Similar to Hypernetworks, we also consider a variation on the LSTM, in part because LSTMs are commonly applied to clinical time-series \citep{fiterau2017shortfuse, lipton2015learning, harutyunyan2017multitask}. 
However, in contrast to previously proposed modifications for handling conditional shift, we impose fewer restrictions on how parameters can be modified at each time step. 

Mixture of experts models are commonly used for multitask learning and conditional computation \citep{kohlmorgen1998analysis, ma2018modeling, wang2018skipnet,eigen2013learning,tan2016cluster,savarese2018learning}.
%A complementary approach to dealing with conditional shift also frames the problem as a multitask problem, but focuses on a mixture of experts.
By framing conditional shift as a multitask problem, we can exploit the large body of work in mixture of experts. \cite{kohlmorgen1998analysis} proposed a two-step approach in which first,  a hidden Markov model (HMM) learns a segmentation of the time series, so that each segment is assigned to an expert, and second, the learned experts are mixed at the segmentation boundaries. 
%More recently, mixture of experts has been investigated in other settings \citep{eigen2013learning,tan2016cluster,savarese2018learning}. 
\cite{eigen2013learning} stacked experts to form a deep mixture of experts; \cite{tan2016cluster} mixed the parameters of fully connected layers, stacking them to account for differences in the training and test sets in audio processing tasks; and \cite{ma2018modeling} learned a gating function to mix the output of experts in a multitask learning setting. The methods proposed by \cite{savarese2018learning} are particularly related. The authors learn coefficients for mixing convolution parameters, increasing parameter sharing across layers of a convolutional neural network (CNN). Building on these approaches, we investigate the utility of a mixture of LSTMs. At each time step, we apply the learned mixing coefficients to form a combined LSTM cell. This facilitates end-to-end learning and allows more than two experts to flexibly contribute to any time step's prediction. Our setting differs from \cite{savarese2018learning} as the mixing coefficients are a) constrained to belong to a simplex, b) learned for each time step instead of each layer, and c) applied to LSTM cells instead of CNN filters.

\section{Methods}\label{sec:methods}

In this section, we describe extensions to LSTMs that facilitate learning in the presence of temporal conditional shift. Building off of an LSTM architecture, we present two variations that relax parameter sharing across time: \texttt{shiftLSTM} and \texttt{mixLSTM}. The first approach, \texttt{shiftLSTM}, represents a simple baseline in which different parameters are learned for different time steps (i.e., separate LSTM cells for different time periods). The second approach, \texttt{mixLSTM}, addresses the shortcomings of this simple baseline through a mixture approach. But first, we formalize the problem setting of temporal conditional shift and review the architecture of an LSTM. 

\subsection{Problem Setup - Temporal Conditional Shift \& LSTMs}
Given time-series data representing patient covariates over time, $\boldsymbol{X}=[ \boldsymbol{x}_1,\boldsymbol{x}_2,...,\boldsymbol{x}_{T}]$ where $\boldsymbol{x}_t\in\mathbb{R}^d$, we consider the task of predicting a sequence of outcomes $\boldsymbol{y}=[y_1,y_2,...,y_{T}]$, where $y_t\in\mathbb{R}$ for $t\in[1,2,3,...,T]$.  We consider a scenario in which the relationship between $\boldsymbol{x}_{1:t}$ and $y_t$ varies over time, i.e., $y_t=f(\boldsymbol{x}_{1:t}, t;\boldsymbol{\theta}_t)$, where $\boldsymbol{\theta}_t$ represent model parameters at time $t$. Because $t$ is measured with respect to a patient-specific fiducial marker, we restrict ourselves to conditional shift within a patient-specific time scale (e.g., within an admission). 

%Unlike online learning where $t$ is defined in terms of global time, here, we calculate $t$ relative to a point that is specific to each patient. For example, in a clinical scenario, $t$ may be measured with respect to the time of admission to the intensive care unit (ICU).

%\subsection{Review of LSTM architecture}

In the sequence-to-sequence setting described above, LSTMs take as input time-varying patient covariates and output a prediction at each time step. Dynamics are captured in part through a cell state $\boldsymbol{C}_t$ that is maintained over time. A standard LSTM cell is described below, where $\ast$ represents element-wise multiplication. Here, $\boldsymbol{h}_t$ and $\tilde{\boldsymbol{C}}_t$ represent the hidden state and the update to the cell state, respectively. 
\begin{align}
    \boldsymbol{i}_t &= \sigma (\boldsymbol{W_i}[\boldsymbol{h}_{t-1},\boldsymbol{x}_t]+\boldsymbol{b_i}) \\
    \tilde{\boldsymbol{C}}_t &= \tanh(\boldsymbol{W_{\tilde{c}}}[\boldsymbol{h}_{t-1},\boldsymbol{x}_t]+\boldsymbol{b_{\tilde{c}}}) \\
    \boldsymbol{f}_t &= \sigma (\boldsymbol{W_f}[\boldsymbol{h}_{t-1},\boldsymbol{x}_t]+\boldsymbol{b_f}) \\
    \boldsymbol{C}_t &= \boldsymbol{i}_t\ast \tilde{\boldsymbol{C}}_t + \boldsymbol{f}_t\ast \boldsymbol{C}_{t-1} \\
    \boldsymbol{o}_t &= \sigma (\boldsymbol{W_o}[\boldsymbol{h}_{t-1},\boldsymbol{x}_t]+\boldsymbol{b_o}) \\
    \boldsymbol{h}_t &= \boldsymbol{o}_t\ast \tanh(\boldsymbol{C}_t) \\
    \hat{y}_t &= \boldsymbol{W}_y \boldsymbol{h}_t + b_y
\end{align}
%refer back to the equation line numbers when describing the new approaches, these equations are here for reference, use it
Importantly, each of the learned parameters $\boldsymbol{W}$ and $\boldsymbol{b}$ in equations (1)-(3), (5) and (7) do not vary with time. To capture time-varying dynamics, the hidden and cell states ($\boldsymbol{h}_t$, $\boldsymbol{C}_t$) must indirectly model conditional shift. 

\vspace{3em}
\setlength{\abovecaptionskip}{3pt plus 3pt minus 2pt} 
\begin{figure}[h]
\floatconts
{fig:overview}% label for whole figure
{\caption{An illustrative plot comparing \texttt{LSTM}, \texttt{shiftLSTM}, and \texttt{mixLSTM}, with four time steps. Each square denotes an LSTM cell. Cells with the same color share the same parameters. Arrows denote transitions between time steps. (a) \texttt{shiftLSTM-$2$} is similar to an LSTM, except it uses different cells for the first two time steps compared to the last two. (b) \texttt{mixLSTM-$2$} has two independent underlying cells, and at each time step, it generates a new cell by mixing a convex combination of the underlying cells. For illustrative purposes, the parameters at each time step are drawn from sequential locations on the continuum, but in reality, the parameter combination is independent of the relative positions of time steps. }}% caption for whole figure
{%
\subfigure[]{%
\label{fig:overview_a}% label for this sub-figure
\includegraphics[width=0.4\linewidth]{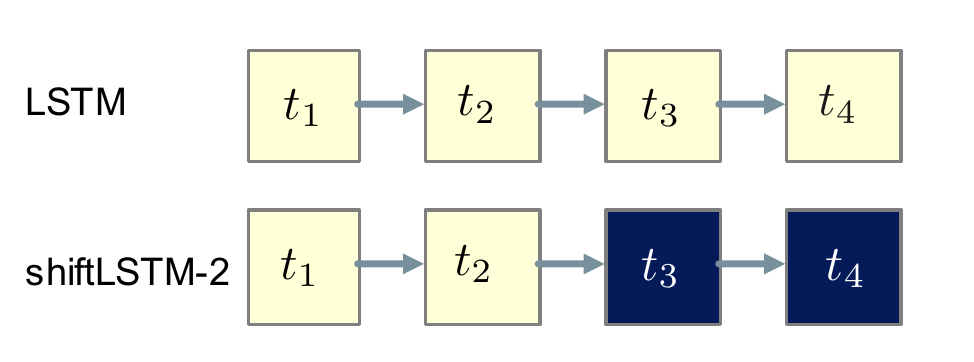}
}\qquad % space out the images a bit
\subfigure[]{%
\label{fig:overview_b}% label for this sub-figure
\includegraphics[width=0.5\linewidth]{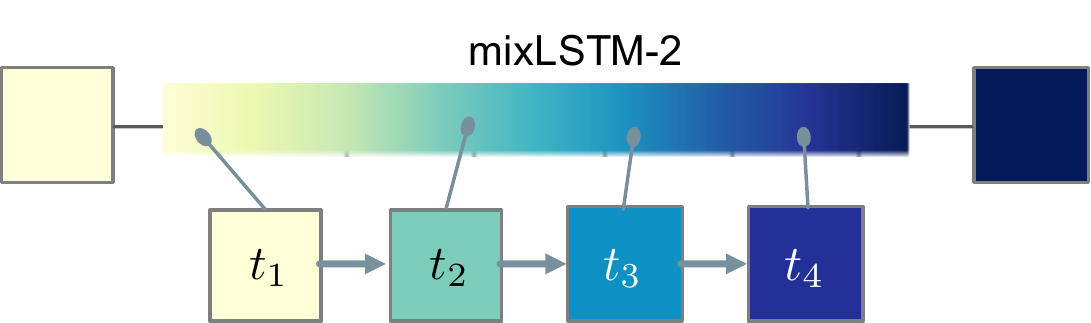}
}
}
\end{figure}

\subsection{Relaxed Parameter Sharing in LSTMs}
We hypothesize that in settings where the amount of training data is limited -- this is often the case in health applications -- an approach that more directly models conditional shift through time-varying parameters will outperform a standard LSTM. To this end, we explore two variations on the LSTM: the \texttt{shiftLSTM} and the \texttt{mixLSTM}, illustrated in \textbf{Figure} \ref{fig:overview}. 

% \begin{figure}[t]
% \centering
%   \includegraphics[width=0.9\linewidth]{overview_approach.pdf} 
%   \caption{An illustrative plot of \texttt{LSTM}, \texttt{shfitLSTM}, and \texttt{mixLSTM}. Each square denotes an LSTM cell. Cells with the same color share the same parameters. Arrows denote transitions between time steps. \texttt{mixLSTM-2} has two independent underlying blocks (base blocks), and at each time step it generates a new cell by mixing a convex combination of the base blocks.} \label{Figure6}
% \end{figure}

\subsubsection{shiftLSTM - learning abrupt transitions}
As a baseline, we consider an approach that na\"ively minimizes parameter sharing across cells, by learning different parameters $\boldsymbol{W}^{(t)}$ and $\boldsymbol{b}^{(t)}$ at each time step $t$, instead of the time-invariant parameters in equations (1)-(3), (5), and (7). This mimics a feed-forward network, with the hidden state and cell state propagating forward at each time step, but computes the output sequentially. This na\"ive approach to relaxed parameter sharing assumes no shared relationships across time. As a result, its capacity is significantly greater than that of an LSTM. Given the same hidden state size, the number of parameters scales linearly with the number of time steps. We hypothesize that this  na\"ive approach will result in overfitting and poor generalization, in settings with limited data. To strike a balance between the two extremes, complete sharing and no sharing, we explore a variation of this approach that assumes parameters are shared across a subset of adjacent time steps: \texttt{shiftLSTM-$K$}. \\

\noindent
\texttt{shiftLSTM-$K$}. This approach sequentially combines $K$ different LSTM cells over time, resulting in different model parameters every $\left \lceil{T/K}\right \rceil$ time steps (\textbf{Figure} \ref{fig:overview_a}). $K\in \{1,...,T\}$ is a hyperparameter, with \texttt{shiftLSTM-$1$} being no different than an LSTM with complete parameter sharing, and \texttt{shiftLSTM-$T$} corresponding to different parameters at each time step. All parameters are learned jointly using backpropagation. %As in a regular LSTM, the model still propagates a hidden state and cell state forward at each time step. 

% \noindent
% \texttt{shiftLSTM-x}. In addition, we considered an approach in which only a subset of gate parameters are allowed to vary over time. We relax parameter sharing in each individual gate (i.e., cell, forget, input, and output) and continue to share in the remaining gates. For example, \texttt{shiftLSTM-o} has shared parameters across all time steps except for time specific output gate parameters $\boldsymbol{W_o^{(t)}}$ and $\boldsymbol{b_o^{(t)}}$.

\subsubsection{mixLSTM - learning smooth transitions}

As described above, the \texttt{shiftLSTM} approach is restricted to sharing parameters within a certain number of contiguous time steps. This not only leads to a substantial increase in the number of parameters, but also results in possibly abrupt transitions. We hypothesize that changes in health data, and risk factors specifically, are gradual. To allow for smooth transitions in time, we propose a mixture-based approach: \texttt{mixLSTM-$K$} (\textbf{Figure} \ref{fig:overview_b}). \\

\noindent
\texttt{mixLSTM-$K$}. Given $K$ independent LSTM cells with the same architecture, let $\boldsymbol{W}^{(k)}$ and $\boldsymbol{b}^{(k)}$ represent the $k^{\text{th}}$ model's weight parameters from equations (1)-(3), (5) and (7). The parameters of the resulting \texttt{mixLSTM} at time step $t$ are
\vspace{-0.25em}
{\small\begin{equation}
\boldsymbol{W}_t = \sum_{k=1}^K \lambda^{(k)}_t \boldsymbol{W}^{(k)}, \qquad \boldsymbol{b}_t = \sum_{k=1}^K \lambda^{(k)}_t \boldsymbol{b}^{(k)}
\end{equation}}\\\vspace{-0.25em}
where $\boldsymbol{\lambda} = \{\lambda_t^{(k)} : t=1\dots T, \ k=1\dots K\}$ are the mixing coefficients and each $\lambda_t^{(k)}$ represents the relevance of the $k^{th}$ model for time step $t$. The mixing coefficients are learnable parameters (initialized randomly) and are constrained such that $\sum_k \lambda_t^{(k)} = 1$ and $\lambda_t^{(k)} \geq 0$. Similar to above, $K$ is also a hyperparameter, but here it can take on any positive integer value. Note that for every $K$, all possible \texttt{shiftLSTM-$K$} models can be learned by \texttt{mixLSTM-$K$}. 

By mixing models, instead of abruptly transitioning from one model to another, \texttt{mixLSTM} can learn to share parameters over time. Moreover, though we do not constrain the mixing coefficients to change smoothly, their continuous nature allows for smooth transitions. We verify these properties in our experiments. 

% \noindent
% \texttt{mixLSTM-output}. Instead of mixing parameters, we can mix the cell outputs at each time step, before passing them to the following time step. 
% %Denote $g_{\boldsymbol{W^{(k)}}}$ as the function parameterized by $\boldsymbol{W^{(k)}}$ so that $g_{\boldsymbol{W^{(k)}}}(\boldsymbol{h_{t-1}}, \boldsymbol{x_t}) = [\boldsymbol{h_t^{(k)}}, \boldsymbol{C_t^{(k)}}]$. Then 
% \begin{align}
% % g_{\boldsymbol{W_t}} (\boldsymbol{h_{t-1}}, \boldsymbol{x_t}) = \sum_{k=1}^K \lambda_t^{(k)} g_{\boldsymbol{W^{(k)}}}(\boldsymbol{h_{t-1}}, \boldsymbol{x_t}) \\
% \boldsymbol{h_t} = \sum_{k=1}^K \lambda_t^{(k)} \boldsymbol{h_t}^{(k)} \\
% \boldsymbol{c_t} = \sum_{k=1}^K \lambda_t^{(k)} \boldsymbol{h_t}^{(k)}
% \end{align}

\section{Experimental Setup}

We explore the effects of temporal conditional shift in both synthetic and real data. Here, we describe i) these datasets, ii) several baselines to which we compare our proposed approach, and iii) the details of our experimental setup. 

\subsection{Synthetic Data}
\label{sec:synth_data}
We begin by considering a scenario in which we can control the extent of conditional shift in the problem. This allows us to test model performance in a setting where the amount of temporal conditional shift is known. Specifically, we consider a multitask variation of the `copy memory task' \citep{arjovsky2016unitary}, with input sequence $\{\boldsymbol{x}_1,\dots,\boldsymbol{x}_T\}$, $\boldsymbol{x}_t\in \mathbb{R}^d$, and output sequence $\{y_{l+1},\dots, y_T\}$, $y_t\in \mathbb{R}$ (we start generating output once we have accumulated $l$ values). The output at each time step is some predetermined, weighted combination of inputs from the previous $l$ time steps, described by two probability vectors, $w^{(l)}_t \in \mathbb{R}^l$ and $w^{(d)}_t \in \mathbb{R}^d$, which are used for weighting the $l$ time steps and $d$ feature dimensions respectively. The parameters change gradually at every time step $t$, such that each time step's weighting (or task) is similar to the task from the previous time step. The parameter $\delta$ controls amount of change between temporally adjacent tasks. The generation process of these parameters is described below, followed by the generation process of the datasets. Here, $\big[\boldsymbol{x}_{t-l}, \dots, \boldsymbol{x}_{t-1}\big]^\intercal \in \mathbb{R}^{l\times d}$ is the concatenation of the previous $l$ inputs at time step $t$. Inputs are generated to be sparse. $\mathtt{Renormalize}(v)$ refers to a renormalization process that ensures the weights in every $w_t$ vector are positive and sum to 1. This is done every time step to ensure that the effect of $\delta$ does not diminish as $t$ increases.  

\scalebox{0.9}{
\begin{minipage}{.5\textwidth}
\begin{algorithm}[H]
\DontPrintSemicolon
\SetKwProg{Pr}{Procedure}{:}{}
\SetKwFunction{FMainb}{SampleWeights}
\Pr{\FMainb{$T$, $l$, $m$}}{
    $w_{l+1} \sim \mathrm{Uniform}(0,1) \in \mathbb{R}^m$ \; \\
    $w_{l+1} = \mathtt{Renormalize}(w_{l+1})$ \; \\
    \For{$t \in \{ l+2, \dots, T\}$}{
        $\Delta_{t} \sim \mathrm{Uniform}(-\delta,\delta) \in \mathbb{R}^{m}$ \; \\
        $w_t = \mathtt{Renormalize}(w_{t-1} + \Delta_t)$ \;
        }
    \KwRet $w_{l+1}, \dots, w_{T}$ \;
} \;
$w_{l+1}^{(d)}, \dots, w_{T}^{(d)}$ = $\mathtt{SampleWeights}(T,l,d)$ \; \\
$w_{l+1}^{(l)}, \dots, w_{T}^{(l)}$ = $\mathtt{SampleWeights}(T,l,l)$ \; \\
\end{algorithm}
\end{minipage}
\hfill
\begin{minipage}{.55\textwidth}
\begin{algorithm}[H]
\DontPrintSemicolon
\SetKwProg{Pr}{Procedure}{:}{}
\SetKwFunction{FMainb}{SampleData}
\Pr{\FMainb{$T$, $w_{l+1:T}^{(l)}$, $w_{l+1:T}^{(d)}$}}{
    \For{$t \in \{1, \dots, T\}$, $i\in\{1,\dots,d\}$}{
        $z_i \sim \mathrm{Bernoulli}(0.1)$ \# for sparse inputs \; \\
        $x_i \sim \mathrm{Uniform}(0,100)$ \; \\
        $x_t[i] = z_i x_i$
        }
    \For{$t \in \{l+1, \dots, T\}$}{
        $y_t = {w^{(l)}_t}^\intercal \big[\boldsymbol{x}_{t-l}, \dots, \boldsymbol{x}_{t-1}\big]^\intercal w^{(d)}_t$ \;
        }
    \KwRet $\{\boldsymbol{x}_1,\dots,\boldsymbol{x}_T\}$, $\{y_{l+1},\dots, y_T\}$ \;
} \;
\vspace{1.2em}
\end{algorithm}
\end{minipage}
}

Our goal is then to learn to predict $\{y_{l+1},\dots, y_T\}$ based on input from the current and all preceding time steps. For each $\delta \in \{0.0, 0.1, 0.2, 0.3, 0.4 \}$, we generated five sets of temporal weights, and then used each set to create five different synthetic dataset tasks where $T=30$, $d=3$, and $l=10$. These twenty-five tasks were kept the same throughout experiments involving synthetic data. Train, validation and test sets all had size $N=1,000$ unless otherwise specified.

\subsection{Clinical Prediction Tasks}\label{clinical_prediction_tasks}

In addition to exploring conditional shift in synthetic data, we sought to test our hypotheses using real clinical data from MIMIC-III \citep{johnson2016mimic}. Below we describe the three clinical prediction tasks of interest, in addition to their corresponding study populations, patient covariates, and evaluation criteria.

\subsubsection{Outcomes}
Throughout the first 48 hours of each ICU visit, we sought to make predictions regarding a patient's risk of experiencing three different outcomes: acute respiratory failure (ARF), shock, and in-hospital mortality, each described in turn below.\\ 

\noindent
\textbf{ARF.} Acute respiratory failure is defined as the need for respiratory support with positive pressure mechanical ventilation \citep{stefan2013epidemiology, meduri1996noninvasive}. Onset time of ARF was determined by either the documented receipt of invasive mechanical ventilation (\texttt{ITEMID: 225792}) or non-invasive mechanical ventilation (\texttt{ITEMID: 225794}) as recorded in the \texttt{PROCEDURESEVENTS\_MV} table, or documentation of positive end-expiratory pressure (PEEP) (\texttt{ITEMID: 220339}) in the \texttt{CHARTEVENTS} table, whichever occurs earlier. Ventilator records and PEEP settings that are explicitly marked as \texttt{ERROR} did not count as an event. \\

\noindent
\textbf{Shock.} Shock is defined as inadequate perfusion of blood oxygen to organs or tissues \citep{gaieski2016shock}, and is characterized by receipt vasopressor therapy. Onset time of shock was determined by the earliest administration of vasopressors \citep{avni2015vasopressors}. Using the \texttt{INPUTEVENTS\_MV} table, we considered the following vasopressors: 
\vspace{-0.5em}
\begin{itemize} \setlength{\itemsep}{-5pt}
\item norepinephrine (\texttt{ITEMID: 221906}), 
\item epinephrine (\texttt{ITEMID: 221289}), 
\item dopamine (\texttt{ITEMID: 221662}), 
\item vasopressin (\texttt{ITEMID: 222315}), and \item phenylephrine (\texttt{ITEMID: 221749}). 
\end{itemize}
\vspace{-0.5em}
\noindent
Drug administration records with the status of \texttt{REWRITTEN}, incorrect units, or non-positive amounts/durations did not count towards an event.\\

\noindent
\textbf{In-hospital mortality.} As in \cite{harutyunyan2017multitask}, the time of in-hospital mortality was determined by comparing patient date of death (\texttt{DOD} column) from the \texttt{PATIENTS} table with hospital admission and discharge times from the \texttt{ADMISSIONS} table.

\subsubsection{Cohort Selection} 
We considered adult admissions with a single, unique ICU visit. This excludes patients with transfers between different ICUs. Patients without labels or observations in the ICU were excluded. Since we are interested in how relationships between covariates and outcome change over time, we focused our analysis on patients who remained in the ICU for at least 48 hours. In addition, for ARF and shock prediction tasks, patients who experienced the event of interest before 48 hours were excluded. Using the full 48 hours allows us to focus on temporal trends that are more likely to be present in longer visits. \textbf{Table} \ref{cohort} shows the number of admissions and positive labels for the three tasks after applying exclusion criteria.

\begin{table}[htbp]    
    \centering 
    \caption{We considered three clinical prediction tasks. The study population varied in size across tasks, as did the fraction of positive cases (i.e., the portion of patients who experienced the outcome of interest.)}
    \scalebox{0.8}{
    \begin{tabular}{cc}
        \toprule
        Task  & {Number of ICU admissions (\%positive)} \\
        \midrule
        ARF  & \,\ 3,789  (\,\ 6.01\%)\\
        shock  & \,\ 5,481  (\,\ 5.98\%)\\
        in-hospital mortality  & 21,139 (13.23\%)\\
        \bottomrule
    \end{tabular}
    }
    \label{cohort}
\end{table}

\subsubsection{Data Extraction and Feature Choices} 

We used the same feature extraction procedure as detailed in \cite{harutyunyan2017multitask}\footnote{\url{https://github.com/YerevaNN/mimic3-benchmarks}}. For completeness, we briefly describe the feature extraction process here. For each ICU admission, we extracted 17 physiological features (e.g., heart rate, respiratory rate, Glasgow coma scale, see \textbf{Table} \ref{tab:17vars} in Appendix \ref{appendix:a}) from the first 48 hours of their ICU visit. We applied mean normalization for continuous values and mapped categorical values to binary features using one-hot encoding, resulting in 59 features. We resampled the time series with a uniform sampling rate of once per hour with carry-forward imputation. Mask features, indicating if a value had been imputed resulted in 17 additional features. After preprocessing, each example was represented by $d=76$ time-series (see \textbf{Table} \ref{tab:76features} in Appendix \ref{appendix:a} for the complete list of features) of length $T=48$ and three binary labels indicating whether or not the patient developed ARF, developed shock or died during the remainder of the hospital stay. 

\subsubsection{Evaluation}

Given these data, the goal was to learn a mapping from the features to a sequence of probabilities for each outcome: ARF, shock or in-hospital mortality. We split the data into training, validation, and test as in \cite{harutyunyan2017multitask}. We used target replication when training the model \citep{lipton2015learning}. For example, if a patient eventually developed ARF, then every hour of the first 48 hours is labeled as positive (negative otherwise). We used the validation set for hyperparameter tuning, and report model performance as evaluated on the held-out test set. Since we consider a sequence-to-sequence setting, each model makes a prediction for every hour during the first 48 hours. These predictions were evaluated based on whether or not at least one prediction exceeds a given threshold. This threshold was swept across all ranges to generate a receiver operating characteristics curve (ROC) and precision-recall curve (PR). This resembles how the model is likely to be used in practice. With the goal of making early predictions, as soon as the real-time risk score exceeds some specified threshold, clinicians could be alerted to a patient's increased risk of the outcome. It should be noted that this differs from the evaluation used in \cite{harutyunyan2017multitask} where a single prediction was made during the 48-hour period. We report performance in terms of the area under the ROC and PR curves (AUROC, AUPR), computing 95\% confidence intervals using 1,000 bootstrapped samples of the test set.

\subsection{Baselines for Comparison}\label{baselines}

In addition to the approaches with relaxed parameter sharing described in the Section \ref{sec:methods}, we considered a number of baselines, which are described below. \\

\noindent
\texttt{NN}. A non-recurrent, feed-forward neural network with one hidden layer. The same network is applied \textit{independently} at every time step to generate the prediction for that time step. This model has complete parameter sharing but no recurrent structure to capture temporal dynamics, and thus is a simpler model than LSTMs and less likely to overfit. This model serves as a simple baseline, but highlights the complexity of the tasks in terms of temporal dynamics. \\

\noindent
\texttt{NN+t}. Similar to \texttt{NN} but with an additional input feature $\hat{x}_t = \frac{t}{T}\in\mathbb{R}$ at every time step, representing the relative temporal position. Given time as an input, this model has the capacity to model temporal conditional shift but cannot leverage longitudinal trends. \\

\noindent
\texttt{LSTM.} We considered a standard LSTM in which parameters are completely shared across time. Synthetic tests used the default Pytorch v0.4.1 implementation (\texttt{torch.nn.LSTM()}). In our experiments on the clinical data, we implemented an LSTM that employed orthogonal parameter initialization and layer normalization, in order to match the settings used in the original HyperLSTM implementation (see below). \\

\noindent
\texttt{LSTM+t}. \texttt{shiftLSTM} and \texttt{mixLSTM} intrinsically have an additional signal regarding the current time step (captured through the use of time-specific parameters). In order to test whether this was driving differences in performance, we tested \texttt{LSTM+t}, an LSTM with an additional input feature $\hat{x}_t = \frac{t}{T}\in\mathbb{R}$ at every time step, representing the relative temporal position.\\

\noindent
\texttt{LSTM+TE}. Given that positional encoding has recently been shown to provide an advantage over simply providing position \citep{vaswani2017attention}, we also explored adding a temporal encoding as additional input features. We used a 24-dimensional encoding for each time step. We tested encoding sizes of 12, 24, 36 and 48 on the in-hospital mortality task, and found 24 to result in the best validation performance. We calculated the temporal encoding as: $TE_{(t,i)} = \sin\left(\frac{t}{10000^{i/24}}\right)$ if $i$ is even, and $ TE_{(t,i)} = \cos\left(\frac{t}{10000^{(i-1)/24}}\right)$ if $i$ is odd, where $t$ represents the time step and $i$ the position in the encoding indexed from $0$. \\

\noindent
\texttt{HyperLSTM}. First proposed by \cite{ha2016hypernetworks}, this approach uses a smaller, auxiliary LSTM to modify the parameters of a larger, primary LSTM at each time step. Since the parameters at each time step are effectively different, this is a form of relaxed parameter sharing. As in the original implementation, we used orthogonal parameter initialization and layer normalization. The two networks were trained jointly using backpropagation. \\

%\texttt{LSTM}, \texttt{LSTM+t}, \texttt{LSTM+TE}, \texttt{HyperLSTM} and \texttt{shiftLSTM} all consisted of single layer recurrent cells that were orthogonally initialized followed by a fully connected layer and a softmax nonlinearity. The hidden state size was hyperparameter tuned for all methods and for \texttt{HyperLSTM} an additional auxiliary hidden state size was optimized. 

\subsection{Model Training \& Implementation Details}

The two non-LSTM baselines both used one hidden layer with ReLU activation and a softmax nonlinearity at the output layer. Except in the case of the LSTM applied to synthetic data, LSTM models consisted of single-layer recurrent cells that were orthogonally initialized followed by a fully-connected layer and a softmax nonlinearity. For experiments involving synthetic data, to compensate for the lower capacity of the LSTM compared to \texttt{mixLSTM} and \texttt{shiftLSTM} which have multiple cells, we allowed it to use an additional layer. Capacity was less of an issue in the experiments involving real data. We tuned the size of the hidden state(s) in all methods based on validation performance.

%For example, the best performing LSTM on the in-hospital mortality task used a hidden state of size only 150. 

%We hypothesize that capacity was more important on the synthetic dataset because of the similarities between the training, validation and test allowed for greater benefits from overfitting. 

We trained all models using the Adam optimizer \citep{kingma2015adam} (Pytorch implementation) with the default learning rate of 0.001. On synthetic data we aimed to minimize the mean squared error (MSE) loss, and for the clinical prediction tasks, we aimed to minimize the cross entropy loss with target replication. We used early stopping based on validation performance -- MSE loss on synthetic data tasks, AUROC on real data tasks -- with a patience of 5 epochs. Models for synthetic datasets were trained with 40 random initializations/hyperparameter settings for a maximum of 30 epochs. We used a batch size of 100 and performed a random search over hidden state sizes of \{100, 150, 300, 500, 700, 900, 1100\}. For learning models on clinical tasks, we used a batch size of 8, because it was the optimal LSTM batch size setting used in the MIMIC-III benchmark paper on the in-hospital mortality task \citep{harutyunyan2017multitask}. When learning models for ARF and shock, we considered 20 random initializations, and trained for a maximum of 30 epochs. For LSTM models, we performed a random search over hidden state and auxiliary hidden states sizes of \{25, 50, 75, 100, 125, 150\}; for \texttt{NN} and \texttt{NN+t}, we performed a random search over the number of hidden units in \{25, 50, \dots, 1000\}. When learning models for in-hospital mortality, we considered 10 random initializations and trained for a maximum of 10 epochs, in part because of the larger training set size. For LSTM models on this task, we performed a random search over hidden state size \{100, 150, 300, 500, 700\} and the same auxiliary hidden state size search as for ARF and shock; for \texttt{NN} and \texttt{NN+t}, we considered the same range of hyperparameters as for ARF and shock. To facilitate comparisons, the code for all of our experiments is publicly available online\footnote{\url{https://gitlab.eecs.umich.edu/MLD3/MLHC2019_Relaxed_Parameter_Sharing}}. 
\vspace{0.5em}

\section{Results \& Discussion} 

In this section, we first show that as temporal conditional shift increases, the performance of the LSTM decreases. Next, we provide evidence that suggests that conditional shift exists in the three clinical prediction tasks. Then, on both the synthetic and real datasets, we show that the proposed method consistently outperforms the baselines. Finally, we present a follow-up analysis focusing on the patterns by which \texttt{mixLSTM} learns to \textit{mix} the parameters, and the robustness of \texttt{mixLSTM} when training data are limited.

\noindent
\subsection{Exploring the Effects of Temporal Conditional Shift}
\begin{figure}[htbp]
\floatconts
{Figure1}% label for whole figure
{\caption{(a.1):  LSTM performance decreases as conditional shift increases. With increasing conditional shift, the time-varying architecture outperforms the LSTM, suggesting that parameter sharing hurts LSTM performance. (a.2) \texttt{mixLSTM} bridges the performance tradeoff between LSTM and \texttt{shiftLSTM}. As conditional shift increases, \texttt{mixLSTM}'s ability to relax parameter sharing helps it increasingly outperform LSTM. By assuming that tasks are unique but related it outperforms \texttt{shiftLSTM}. (b) This issue is only apparent when training data are limited; LSTMs can adapt to temporal conditional shift given enough training data. Error bars represent 95\% confidence intervals based on bootstrapped samples of the test set. $\delta$ was set to $0.3$.}}% caption for whole figure
{%
%\subfigure[Performance of LSTM, \texttt{shiftLSTM-$30$}, and \texttt{mixLSTM-$2$} on synthetic data with varying amounts of temporal conditional shift (increases with $\delta$).]{%
\subfigure[]{%
\label{Figure1a}% label for this sub-figure
\includegraphics[width=0.4\linewidth]{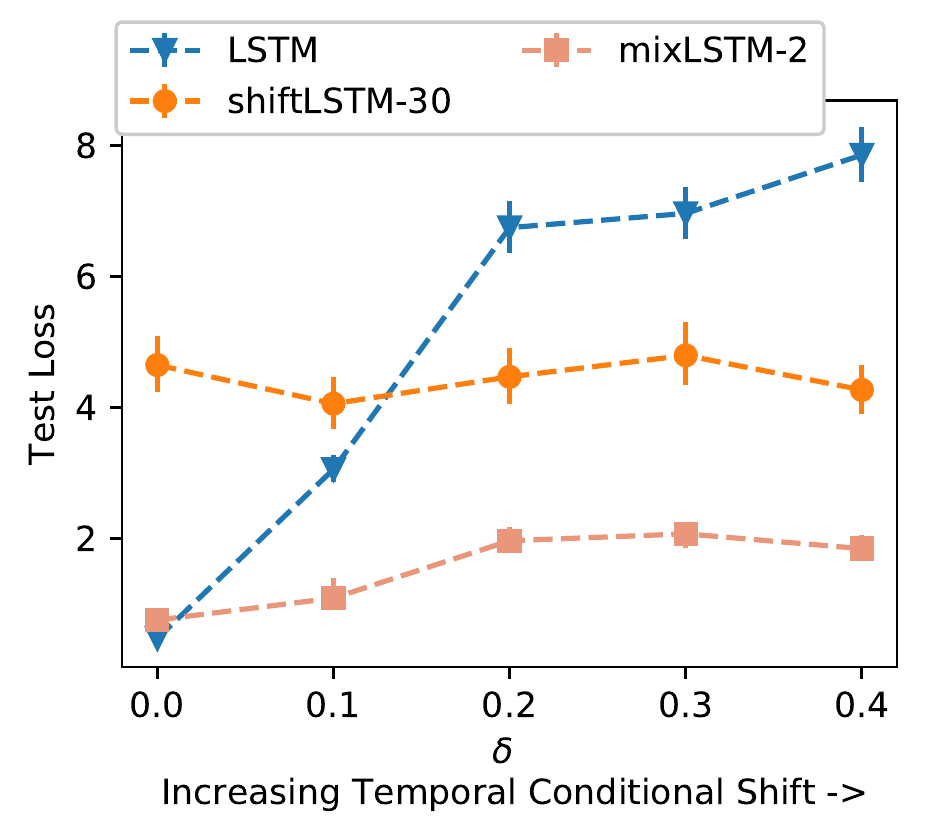}
}\qquad % space out the images a bit
%\subfigure[LSTM performance on synthetic data generated at $\delta=0.3$ over varying amounts of training data.]{%
\subfigure[]{%
\label{Figure1b}% label for this sub-figure
\includegraphics[width=0.4\linewidth]{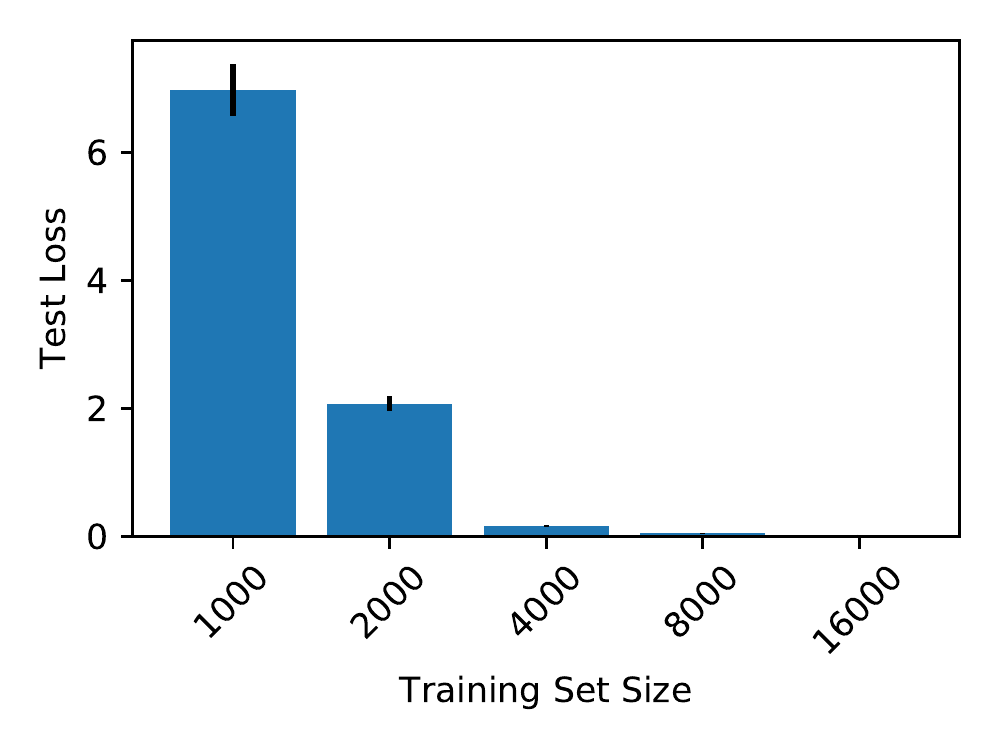}
}
}
\end{figure}

\textbf{Does parameter sharing hinder the ability of an LSTM to capture time-varying relationships?}
The data generation process described in Section \ref{sec:synth_data} allows us to control the amount of temporal conditional shift present in the task. Specifically, by increasing $\delta$, we increase the variability between two temporally adjacent tasks. This allows us to test the effects of conditional shift on the performance of an LSTM. We hypothesize that because the LSTM shares parameters over time, it will struggle to adapt to temporal conditional shift. To test our hypothesis, we compare the performance of an LSTM with \texttt{shiftLSTM} across a range of $\delta$ values (\textbf{Figure} \ref{Figure1a}). Here, the \texttt{shiftLSTM} approach learns different sets of parameters for each time step (30 in total). We observe a clear trend: as temporal conditional shift increases, the performance of the LSTM decreases. In contrast, \texttt{shiftLSTM} results in steady performance across the range of $\delta$. At low $\delta \in \{0,0.1\}$, the LSTM outperforms the \texttt{shiftLSTM} in terms of MSE on the test set. In this experiment, we limited the amount of training data to 1,000 samples. Theoretically, given enough training data, LSTM should be capable of accurately modeling  time-varying relationships. To verify this,  we show that the test loss associated with the LSTM models approaches zero as the training set size increases (\textbf{Figure} \ref{Figure1b}). These results support our initial hypothesis that in settings with limited data, temporal conditional shift negatively impacts LSTM performance and that this impact is in part due to the sharing of parameters. \\

\noindent
\textbf{Is there any evidence of time-varying relationships in the three clinical prediction tasks of interest?}
We tested for the presence of temporal conditional shift in three clinical prediction tasks: ARF, shock, and in-hospital mortality. For these tasks, the underlying parameters that govern the amount of temporal conditional shift (e.g., $\delta$) are unknown. Instead, we indirectly measure temporal conditional shift by applying \texttt{shiftLSTM-$K$} varying $K$ from $\{1, 2, 3, 4, 8, 48\}$, where $K=1$ is a standard LSTM, and $K=48$ implies a different set of parameters for each time step.  Increasing the number of cells or $K$ reduces sharing. As the difference between sequential tasks increases, we expect the benefit of learning different LSTM cells (less parameter sharing) to increase. Empirically, we observe that less parameter sharing results in better performance (\textbf{Figure} \ref{Figure2}). This supports our hypothesis that architectures for solving clinical prediction tasks could benefit from relaxed parameter sharing.  

\begin{figure}[htbp]
\floatconts
{Figure2}% label for whole figure
{\caption{As we increase parameter sharing by increasing $T/K$ along the x-axis, there is a drop in performance, supporting our hypothesis that temporal conditional shift is present in our real data tasks. Error bars represent the interquartile ranges based on bootstrapped samples of the test set.}}% caption for whole figure
{%
\subfigure[ARF]{%
\label{Figure2ARF}% label for this sub-figure
\includegraphics[width=0.32\linewidth]{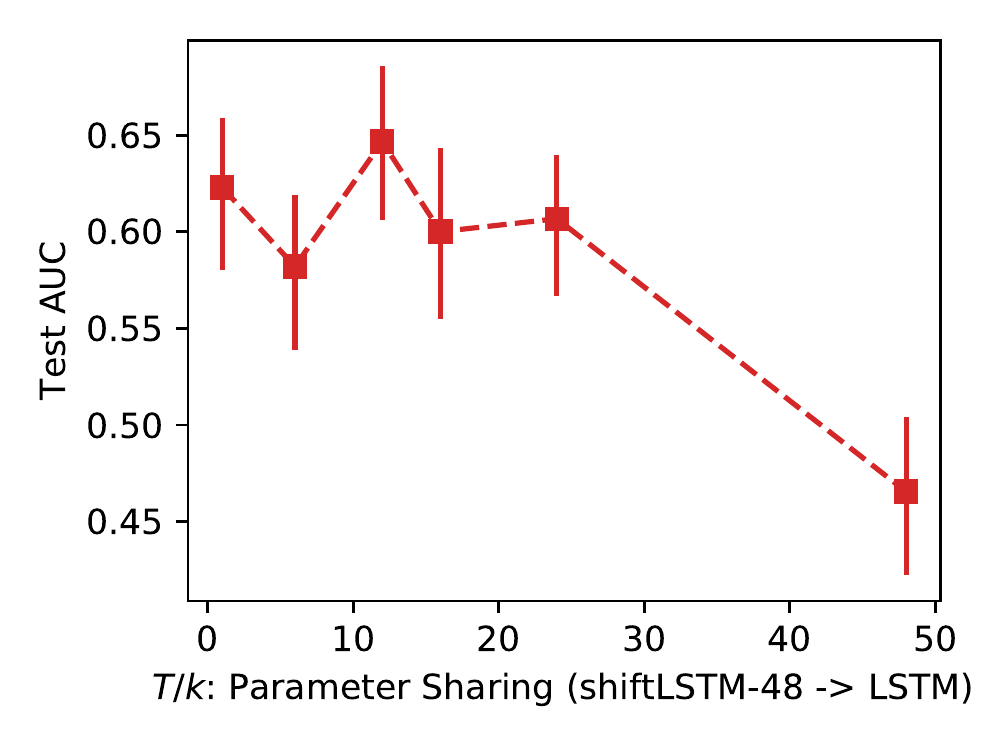}
}%\qquad % space out the images a bit
\subfigure[shock]{%
\label{Figure2Shock}% label for this sub-figure
\includegraphics[width=0.32\linewidth]{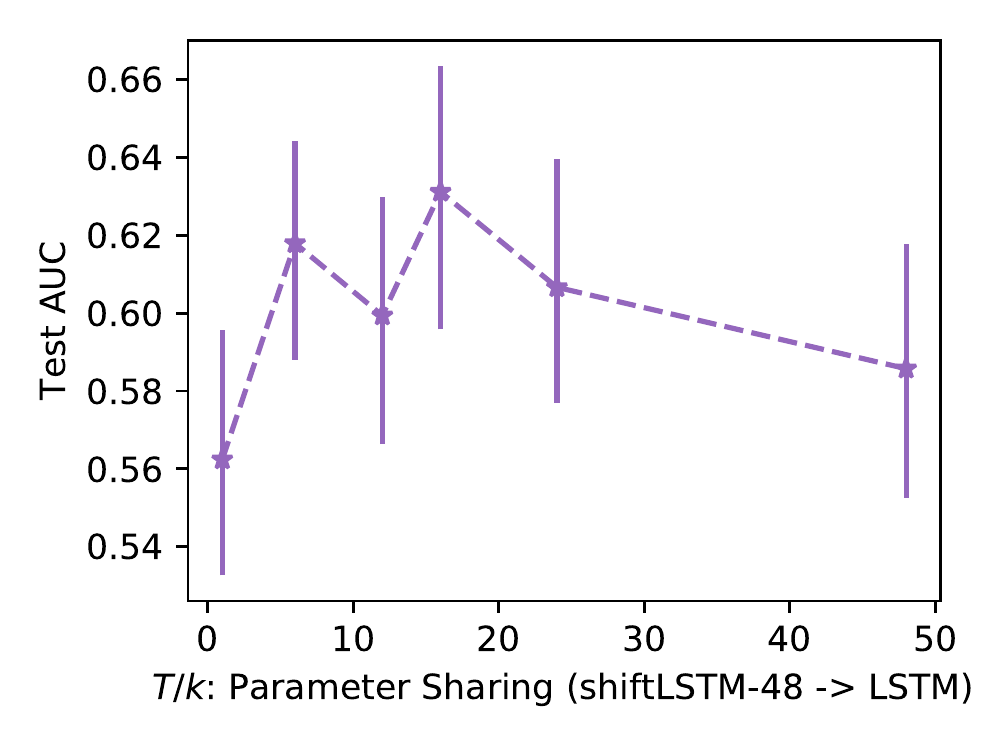}
}%\qquad % space out the images a bit
\subfigure[mortality]{%
\label{Figure2mortality}% label for this sub-figure
\includegraphics[width=0.32\linewidth]{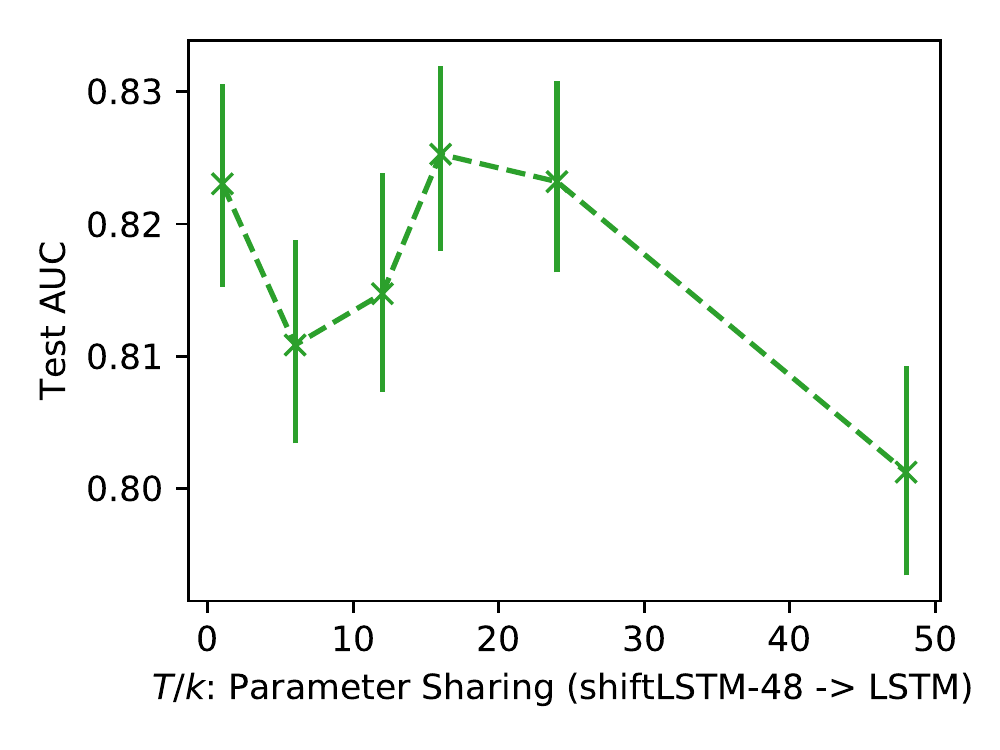}
}
}
\end{figure}

\vspace{-1em}
\subsection{Comparing the Proposed Approach to Baselines}

In this section, we explore the performance of the proposed approach, \texttt{mixLSTM}, relative to the other baselines. Again, we hypothesize that it will outperform the other approaches due to a) smooth sharing of parameters and b) the ability to learn which cells to share. \texttt{mixLSTM} strikes a balance between complete parameter sharing (LSTM) and no parameter sharing (\texttt{shiftLSTM-$48$}). In addition, compared to \texttt{shiftLSTM}, \texttt{mixLSTM} can share parameters between distant time steps and learns how to accomplish this.\\ 

\noindent
\textbf{How does the proposed approach perform on synthetic data?} \texttt{mixLSTM} has the ability to continuously interpolate between $K$ independent cell parameters. In this instance, \texttt{mixLSTM-$2$} has 15 times fewer parameters relative to \texttt{shiftLSTM-$30$}. 
%Therefore, we hypothesize that it will i) outperform LSTM in the presence of temporal conditional shift and ii)  outperform \texttt{shiftLSTM} when training data are limited. 
On the synthetic data tasks, \texttt{mixLSTM} consistently outperforms \texttt{shiftLSTM} at all levels of temporal shift (\textbf{Figure} \ref{Figure1a}). Moreover, \texttt{mixLSTM} outperforms LSTM at low $\delta$, except when no temporal shift exist. This agrees with our intuition that \textit{smart} sharing is better than no sharing (\texttt{shiftLSTM}) and indiscriminate sharing (LSTM). \\

\begin{table}[htbp]
    \renewrobustcmd{\bfseries}{\fontseries{b}\selectfont}
    \centering 
    \caption{Performance on ARF, shock, \& mortality with 95\% confidence intervals. Though the differences are small, \texttt{mixLSTM} consistently outperforms the other approaches across all tasks in terms of both AUROC and AUPR. The number of test samples for each task is reported in parentheses.}
    \scalebox{0.72}{
    \hskip-0.0cm\begin{tabular}{lcccccc}
    %  &  \multicolumn{3}{c}  &  \multicolumn{3}{c} {AUC-PR [95\% CI]}\\
    \toprule
            &  \multicolumn{2}{c}{\textbf{ARF}} & \multicolumn{2}{c}{\textbf{shock}} &  \multicolumn{2}{c}{\textbf{mortality}}\\ 
{Model}     &  \multicolumn{2}{c}{(n=549)}      & \multicolumn{2}{c}{ (n=786) }      &  \multicolumn{2}{c}{(n=3,236)}\\
    
     & {AUROC} & {AUPR} & {AUROC} & {AUPR} &{AUROC} & {AUPR} \\
    \midrule
    \texttt{NN}         & 0.57 [0.45, 0.67]  & 0.05 [0.03, 0.10] & 0.60 [0.52, 0.68] & 0.08 [0.05, 0.12] & 0.80 [0.77, 0.82] & 0.39 [0.34, 0.44] \\
    \texttt{NN+t}       & 0.58 [0.46, 0.68]  & 0.06 [0.03, 0.11] & 0.56 [0.47, 0.64] & 0.10 [0.05, 0.19] & 0.79 [0.77, 0.82] & 0.38 [0.33, 0.43] \\
    \texttt{LSTM}       & 0.47 [0.35, 0.58]  & 0.04 [0.02, 0.07] & 0.59 [0.49, 0.69] & 0.09 [0.05, 0.16] & 0.80 [0.78, 0.83] & 0.39 [0.33, 0.43] \\
    \texttt{LSTM+t}     & 0.42 [0.30, 0.54]  & 0.04 [0.02, 0.07] & 0.62 [0.53, 0.70] & 0.08 [0.05, 0.15] & 0.81 [0.79, 0.83] & 0.41 [0.36, 0.47] \\
    \texttt{LSTM+TE}    & 0.48 [0.35, 0.61]  & 0.05 [0.03, 0.10] & 0.60 [0.50, 0.69] & 0.10 [0.06, 0.20] & 0.82 [0.80, 0.85] & 0.43 [0.38, 0.48] \\
    \texttt{HyperLSTM}  & 0.57 [0.44, 0.68]  & 0.06 [0.03, 0.10] & 0.63 [0.54, 0.72] & 0.08 [0.05, 0.12] & 0.82 [0.80, 0.84] & 0.42 [0.37, 0.47] \\
    \texttt{shiftLSTM}  & 0.61 [0.49, 0.70]  & 0.10 [0.03, 0.21] & 0.61 [0.52, 0.70] & 0.09 [0.05, 0.16] & 0.81 [0.79, 0.84] & 0.43 [0.37, 0.48] \\
    \texttt{mixLSTM}    & {\bfseries 0.72} [0.62, 0.80] & {\bfseries 0.15} [0.06, 0.27] &  {\bfseries 0.67} [0.58, 0.76]  & {\bfseries 0.10} [0.06, 0.16]  &  {\bfseries 0.83} [0.81, 0.85] & {\bfseries 0.45} [0.40, 0.50] \\
    \bottomrule
    \end{tabular}}
    \label{Table1}
    \vspace{-1.2em}
\end{table}

\noindent
\textbf{How does the proposed approach perform on the clinical prediction tasks?} Applied to the three clinical prediction tasks (with varying amounts of training data), \texttt{mixLSTM} consistently performs the best (\textbf{Table} \ref{Table1}). The \texttt{NN} and \texttt{NN+t} models are simpler architectures that outperform other LSTM-based baselines only under very low data settings (ARF). Compared to the LSTM baseline, \texttt{LSTM+t} and \texttt{LSTM+TE} performed better given sufficient training data, suggesting that having direct access to time either as a feature or a temporal encoding is beneficial. Relaxing parameter sharing further improves performance. As shown earlier, \texttt{shiftLSTM} consistently improves performance over the standard LSTM. 

HyperLSTM, similar to \texttt{mixLSTM}, bridges the dichotomy of completely shared and completely independent parameters, and outperforms both LSTM and \texttt{shiftLSTM} in some cases but not consistently. \texttt{mixLSTM} outperforms all other baselines on all three tasks, though the differences are not statistically significant in all cases. Both HyperLSTM and \texttt{mixLSTM} achieve high performance and both models relax parameter sharing. This supports our hypothesis that relaxed parameter sharing is beneficial in some settings. 

In these experiments, we selected $K$ for each task based on validation performance, testing $K\in \{2,3,4,8,48\}$ for \texttt{shiftLSTM} and sweeping $K$ from $2$ to $4$ for \texttt{mixLSTM}. For \texttt{shiftLSTM}, $K$ represents the optimal number of sequential tasks to segment the input sequence into; the best $K=2,2,8$ for ARF, shock, and mortality respectively. For \texttt{mixLSTM}, $K$ indicates the optimal number of operational or characteristic modes in the data; the best $K=4,4,2$, respectively. It appears that for \texttt{shiftLSTM}, the chosen $K$ is correlated with the amount of training data available. Both ARF and shock have significantly smaller training set sizes compared to mortality. In contrast, \texttt{mixLSTM} learns more cells for ARF and shock. This suggests that the structure of \texttt{mixLSTM} is better suited to the problem setting than \texttt{shiftLSTM}, since it is able to train twice as many cells as \texttt{shiftLSTM} and attain a higher test performance. The converse also supports this claim. \texttt{mixLSTM} is able to train $\frac{1}{4}$ the number of cells as \texttt{shiftLSTM} for mortality and still attain better performance. The optimal $K$ for \texttt{mixLSTM} appears to be less indicative of training set size, and more a reflection of the true number of operational or characteristic modes in the data. When we visualize the mixing ratios learned by \texttt{mixLSTM-2} in later sections (\textbf{Figure} \ref{fig:vis_mixWeight}) we see that while mortality smoothly interpolates between \textsf{cell1} and \textsf{cell2} as time passes, ARF and shock both display an initial peak followed by a gradual interpolation. This suggests that the dynamics are more complex for ARF and shock.  

%The difference in optimal $K$ values between \texttt{shiftLSTM} and \texttt{mixLSTM} present interesting complementary information. For \texttt{shiftLSTM}, $K$ represents the optimal number of sequential tasks to segment the input sequence into. For \texttt{mixLSTM}, $K$ indicates the optimal number of operational or characteristic modes in the data. Therefore differences in optimal $K$ values between the two methods present an interesting picture about the nature of the time varying relationships in the data. 

\subsection{Robustness and Sensitivity Analyses}

In this section, we further analyze \texttt{mixLSTM}, focusing on its robustness in settings when training data are limited and investigate what it has learned in terms of mixing trends and changing feature importance. 

\subsubsection{Performance with Limited Training Data}
\begin{figure}[htbp]
\centering
  \includegraphics[width=0.40\linewidth]{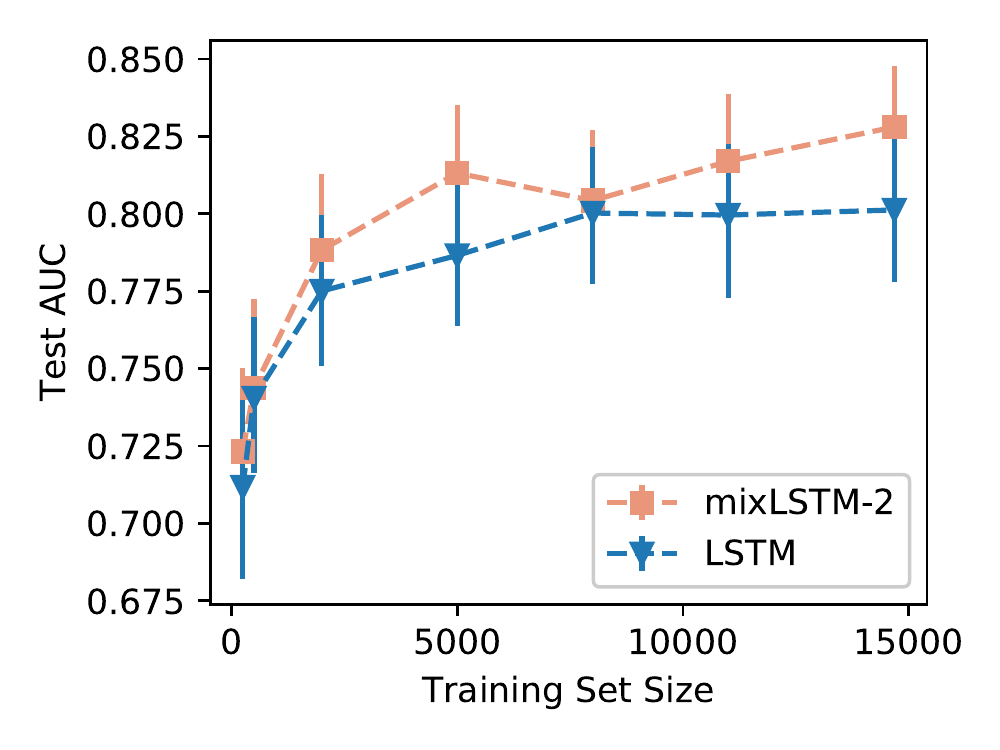} 
  \caption{\texttt{mixLSTM-$2$} is consistently better than LSTM at different training set sizes. Error bars represent 95\% confidence intervals bootstrapped from the test set.   }\label{Figure6}
\end{figure}

\textbf{Does the proposed approach still perform well when  training data are limited?} We hypothesized that \texttt{mixLSTM} will continue to outperform LSTM, even when training data are limited because \texttt{mixLSTM}s are better suited to problem settings exhibiting temporal conditional shift. To test our hypothesis, we compared the performance of \texttt{mixLSTM-$2$} and LSTM trained using different training set sizes for the task of predicting in-hospital mortality. We chose to focus on the task of in-hospital mortality, since it had the most training data (training set size $=14,681$). We subsampled the training set repeatedly for $N \in \{250, 500, 2000, 5000, 8000, 11000\}$. The test set was held constant across all experiments and $K=2$ to limit the capacity of the model. \texttt{mixLSTM} consistently outperforms LSTM across all ranges of training set sizes (\textbf{Figure \ref{Figure6}}). As one might expect, differences are subtle at smaller training set sizes, where an LSTM with complete parameter sharing is likely more sample efficient and therefore more competitive.

\subsubsection{What has mixLSTM learned?}

To dive deeper into what exactly the \texttt{mixLSTM} has learned, we visualize the learned mixing coefficients and the most important features.\\

\begin{figure}[htbp]
  \centering
    \includegraphics[width=0.9\textwidth]{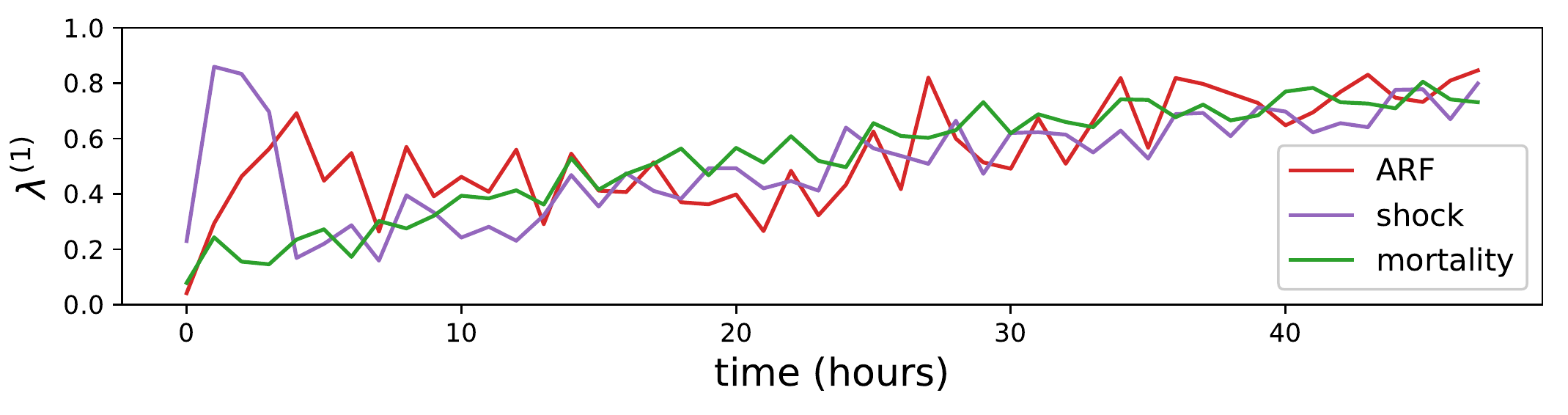}
  \caption{Visualization of the mixing coefficients learned by \texttt{mixLSTM-2}. $\lambda^{(1)}$ is shown on the y-axis, while $\lambda^{(2)}$ can be inferred ($\lambda^{(2)} = 1 - \lambda^{(1)}$). Although not constrained to be smooth, we observe a smooth transition of mixing coefficients between time steps, indicating that one cell is specialized for the beginning of a patient's ICU stay while the other is specialized for 48 hours into the ICU stay.}
\label{fig:vis_mixWeight}
\end{figure}

\noindent
\textbf{Are \texttt{mixLSTM}'s learned mixing coefficients smooth?}
In our learning objective function, \texttt{mixLSTM}'s mixing coefficients are not constrained to be smooth. However, we hypothesize that this behavior reflects the underlying dynamics in clinical data. \textbf{Figure} \ref{fig:vis_mixWeight} plots the mixing coefficients ($\lambda^{(1)}$) over time for \texttt{mixLSTM-2} on the three clinical prediction tasks. Since there are only two independent cells ($K=2$), we can infer $\lambda^{(2)} = 1 - \lambda^{(1)}$. The trend indicates that one cell captures the dynamics associated with the beginning of a patient's stay, while the second cell captures the dynamics 48 hours into the stay. \\

\noindent\textbf{Does explicitly smoothing the mixing coefficients help?}
Based on patterns displayed in \textbf{Figure} \ref{fig:vis_mixWeight}, we hypothesized that additional smoothing of the mixing coefficients could aid classification performance. To test this hypothesis, we applied regularization based on a similarity measure between models at consecutive time steps. Following \cite{savarese2018learning}, we used the normalized cosine similarity as the similarity measure. For consecutive time steps $t$ and $t+1$, this similarity score is $s_t = \frac{\langle \boldsymbol{\lambda}_t, \boldsymbol{\lambda}_{t+1}\rangle}{\| \boldsymbol{\lambda}_t \|_2 \| \boldsymbol{\lambda}_{t+1} \|_2}$ where $\boldsymbol{\lambda}_t := [\lambda^{(1)}_t, \cdots, \lambda^{(K)}_t]$. Denoting $\mathcal{L}$ as the original loss function and $\alpha \in \mathbb{R}^+$ as the regularization strength, we minimize the regularized objective $\mathcal{L}_R := \mathcal{L} - \alpha \sum_{t=1}^{T-1} s_t$ to encourage temporal smoothness.

\begin{figure}[htbp]
  \centering
    \includegraphics[width=0.9\textwidth]{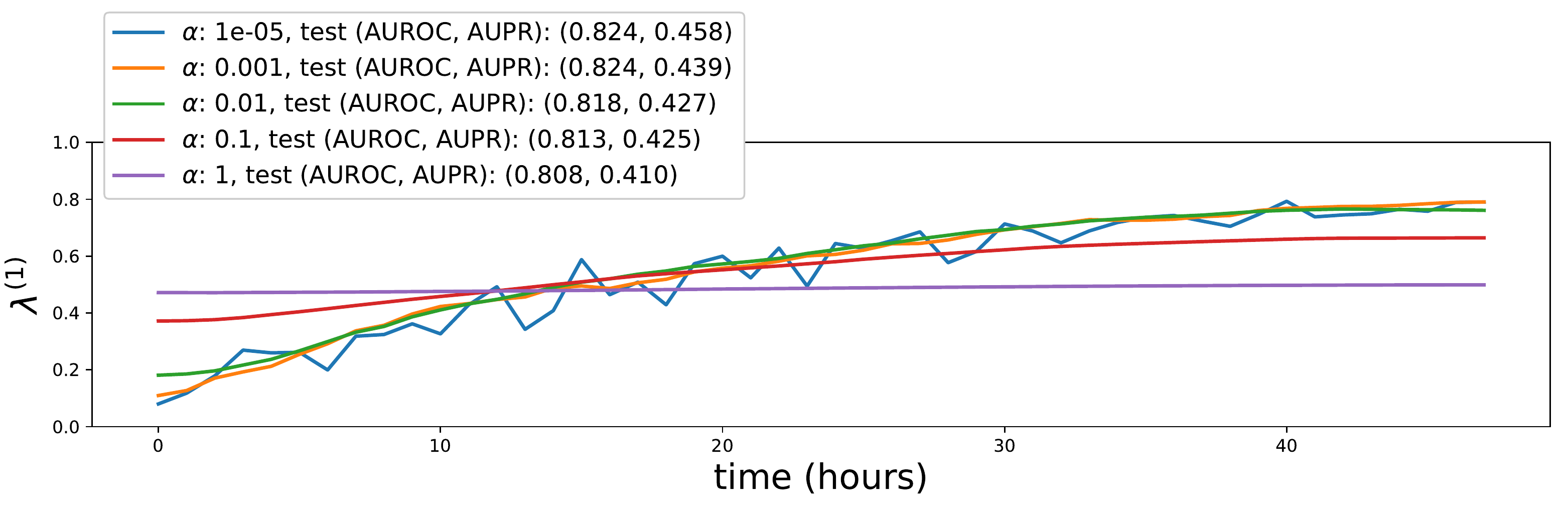}
  \caption{Visualization of the mixing coefficients learned by \texttt{mixLSTM-2} with different regularization strengths for the mortality task. $\lambda^{(1)}$ is shown on the y-axis. $\lambda^{(2)}$ can be inferred ($\lambda^{(2)} = 1 - \lambda^{(1)}$). As regularization strength increase, mixing coefficients become smoother at the cost of lower performance.}
\label{fig:vis_mixWeight_ihm}
\end{figure}

\textbf{Figure} \ref{fig:vis_mixWeight_ihm} illustrates the effect of temporal smoothness regularization on the model for the mortality task. As expected, larger regularization strength encourages models to share parameters. However, test performance drops monotonically as $\alpha$ increases. Additional regularization likely results in lower model complexity and in some settings underfitting. Our result aligns with \cite{savarese2018learning} in that while smoothness in patterns naturally arises, explicitly encouraging smoothness through regularization hurts performance. \\%thus it comes as no surprise that temporal smoothness regularization does not help.

\begin{figure}[ht]
\floatconts
{fig:saliency}% label for whole figure
{\caption{Input gradient based saliency map of \texttt{mixLSTM-$2$} on three tasks. Each plot shows a proxy of importance of each feature across time steps. Some noticeable temporal patterns include high variability during the first six hours, which may be a reflection of increased physiological change a patient may experience at the beginning of their ICU stay when interventions are more frequent.}}% caption for whole figure
{%
\subfigure[ARF]{%
\label{fig:saliencyARF}% label for this sub-figure
\includegraphics[width=0.31\linewidth]{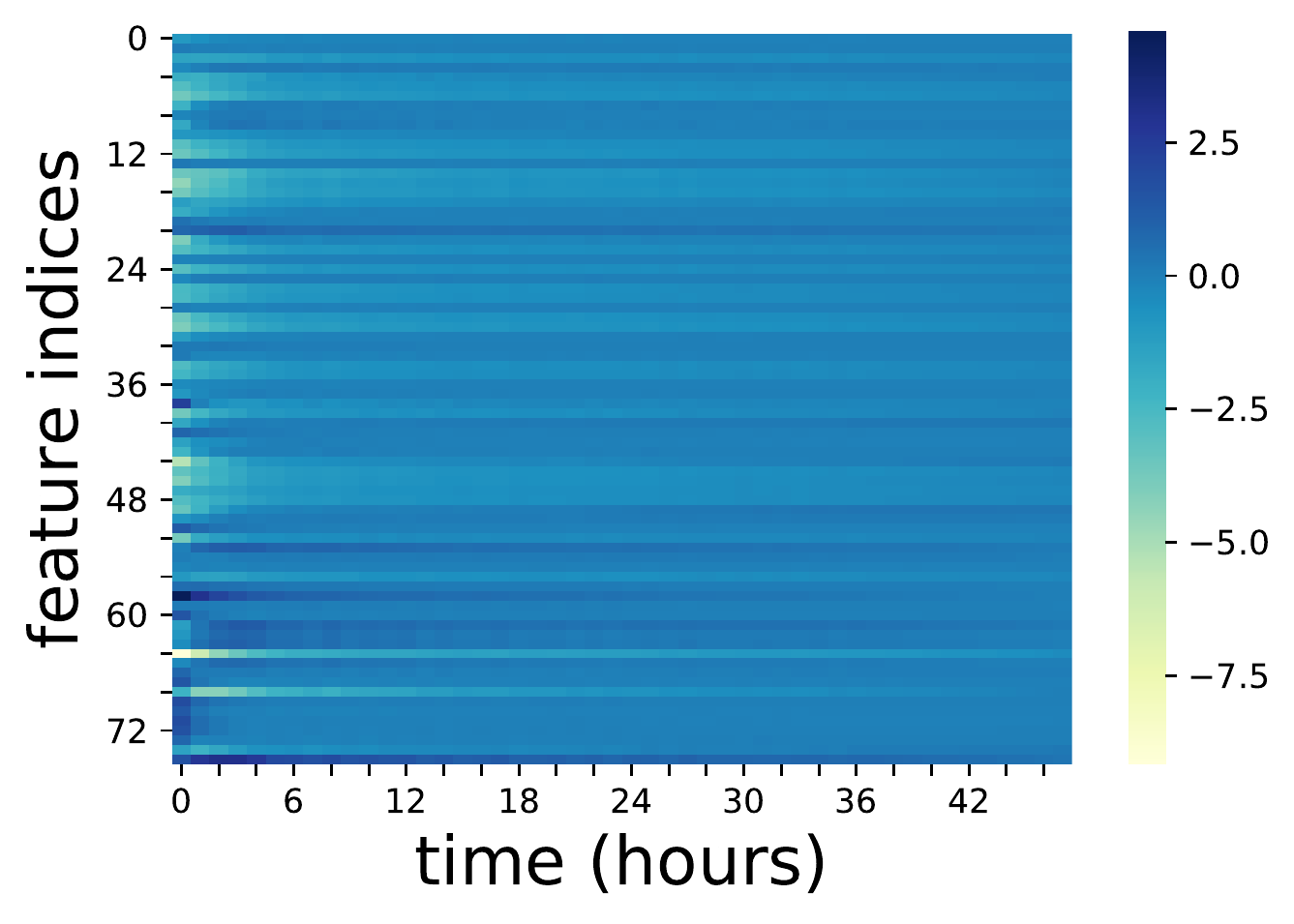}
}%\qquad % space out the images a bit
\subfigure[shock]{%
\label{fig:saliencyShock}% label for this sub-figure
\includegraphics[width=0.31\linewidth]{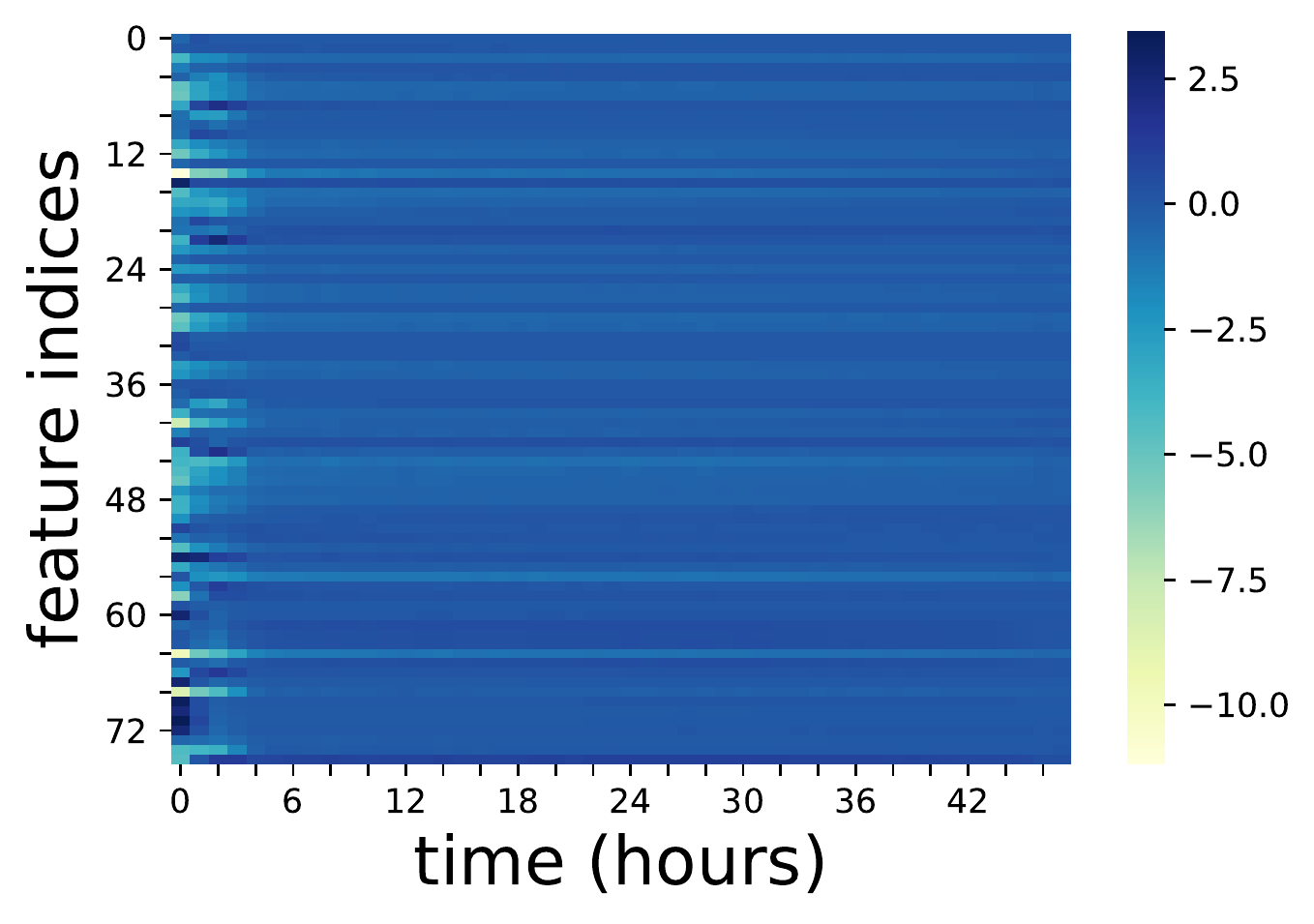}
}%\qquad % space out the images a bit
\subfigure[mortality]{%
\label{fig:saliencymortality}% label for this sub-figure
\includegraphics[width=0.31\linewidth]{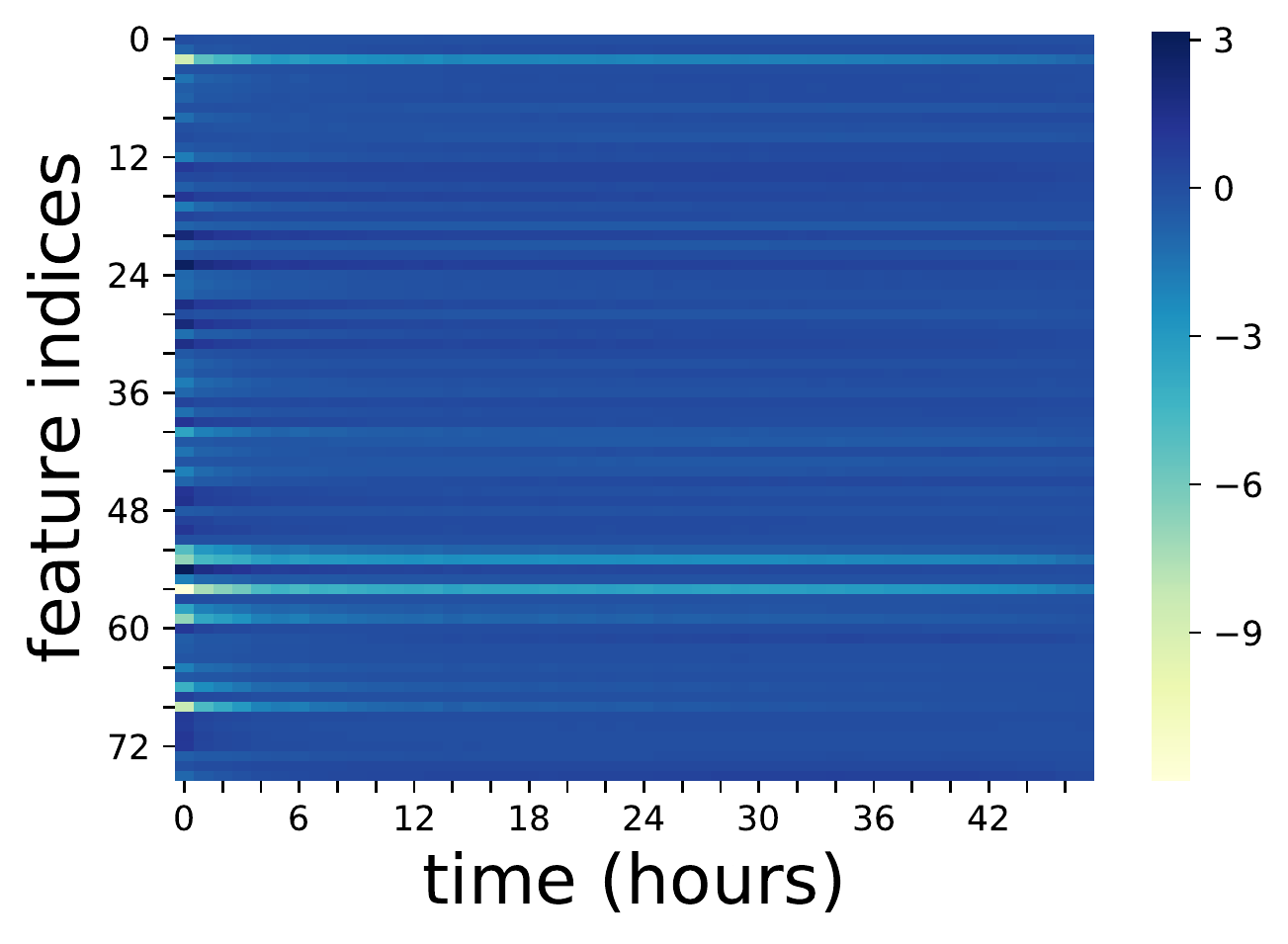}
}
}
\end{figure}

\noindent
\textbf{What time-varying relationships does the \texttt{mixLSTM} learn to recognize?} When attempting to understand which features drive a model's predictions, the focus is often put on the importance of certain features. However, because \texttt{mixLSTM} was designed to and has been shown to excel in situations with temporal conditional shift, we focus on identifying the features whose influence changes over time. To identify such features, we must first measure the effect of each feature at each time step. Here, we use the input gradient as a proxy for feature importance and visualize importance over time \citep{van2016deep, selvaraju2016grad, graves2012supervised} (\textbf{Figure} \ref{fig:saliency}). More specifically, we traversed the test set, accumulating the input gradient with respect to the target class. One of the most noticeable patterns is the large amount of variation in feature importance in the first 6 hours of an ICU admission. This pattern is most apparent for the task of predicting shock (\textbf{Figure} \ref{fig:saliencyShock}). This may reflect the significant physiological changes a patient may experience at the beginning of their ICU stay as interventions are administered in an effort to stabilize them.

\definecolor{myPink}{rgb}{0.949,0.647,0.694}
\definecolor{myRed}{rgb}{0.690,0.137,0.094}
\definecolor{myLime}{rgb}{0.744,0.854,0.630}
\definecolor{myGreen}{rgb}{0.193,0.279,0.111}
\begin{table}[ht]    
    \centering 
    \caption{Physiological data ranked by overall importance as identified by \texttt{mixLSTM-$2$} on ARF, shock, and mortality using input gradient. The table is color coded. Light red denotes features that are initially risk factors, where risk \textit{decreases} over time. Dark red denotes features that are initially risk factors, but where risk \textit{increases} over time. Light green denotes features that are initially protective, but become less protective over time. Dark green denotes features that are initially protective, and becomes more protective over time. }
    \scalebox{0.75}{
    \begin{tabular}{p{4.5cm}p{4.5cm}p{4.5cm}} 
        \toprule
        {\textbf{ARF}} & {\textbf{shock}} & {\textbf{mortality}}\\ \midrule
\cellcolor{myPink}pH & \cellcolor{myPink}Respiratory rate & \cellcolor{myPink}Respiratory rate\\
\cellcolor{myRed}\textcolor{white}{Oxygen saturation} & \cellcolor{myPink}Height & \cellcolor{myPink}Heart Rate\\
\cellcolor{myPink}Weight & \cellcolor{myRed}\textcolor{white}{Mean blood pressure} & \cellcolor{myPink}Glucose\\
\cellcolor{myPink}Respiratory rate & \cellcolor{myRed}\textcolor{white}{Heart Rate} & \cellcolor{myRed}\textcolor{white}{Fraction inspired oxygen}\\
\cellcolor{myRed}\textcolor{white}{Fraction inspired oxygen} & \cellcolor{myRed}\textcolor{white}{Fraction inspired oxygen} & \cellcolor{myLime}Height\\
\cellcolor{myRed}\textcolor{white}{Heart Rate} & \cellcolor{myRed}\textcolor{white}{pH} & \cellcolor{myGreen}\textcolor{white}{Weight}\\
\cellcolor{myPink}Height & \cellcolor{myRed}\textcolor{white}{Weight} & \cellcolor{myLime}Systolic blood pressure\\
\cellcolor{myLime}Glucose & \cellcolor{myLime}Glucose & \cellcolor{myLime}pH\\
\cellcolor{myLime}Systolic blood pressure & \cellcolor{myLime}Oxygen saturation & \cellcolor{myLime}Mean blood pressure\\
\cellcolor{myLime}Mean blood pressure & \cellcolor{myLime}Systolic blood pressure & \cellcolor{myLime}Diastolic blood pressure\\
\cellcolor{myLime}Temperature & \cellcolor{myLime}Diastolic blood pressure & \cellcolor{myLime}Oxygen saturation\\
\cellcolor{myLime}Diastolic blood pressure & \cellcolor{myGreen}\textcolor{white}{Temperature} & \cellcolor{myLime}Temperature\\         \bottomrule
    \end{tabular}
    }
    \label{table:important_factors_continuous}
\end{table}

We list the continuous features ranked by importance in \textbf{Table} \ref{table:important_factors_continuous}. Feature importance was calculated by summing importance over time and taking the absolute value. Here positive importance values are associated with increased risk, while negative importance values are associated with protection. The color scheme reflects the overall direction of association and the change over time. Dark red and dark green represent `risk' and `protective' factors that lead to increased and decreased risk over time, respectively; that is, their effects become amplified over time. Light red and light green represent `risk' and `protective' factors that lead to decreased and increased risk over time, respectively; that is, their effects diminish over time. For example, in all three tasks, `fraction of inspired oxygen' (indicative of whether or not a patient is on supplemental oxygen) is a risk factor initially, and becomes more important over time. This suggests that if a patient is still on high levels of oxygen 48 hours into their ICU admission, their risk is elevated for all three outcomes. For ARF and shock a similar pattern holds for heart rate, where sustained high heart rate is associated with greater risk over time. This suggests that some features, when persistently abnormal, further amplify a patient's risk.

%For in-hospital mortality, temperature is initially protective, a temperature that remains high is associated with increased risk. This finding agrees at least in part with recent work identifying an association between the trajectory of temperature and adverse outcomes \citep{bhavani2019identifying}.

It is important to note that interpreting neural networks, and LSTMs in particular, remains an open challenge. Though the approach considered here is frequently used for interpreting LSTMs, it relies on the local effect of a feature and thus ignores the global trends \citep{ross2017right, ghorbani2017interpretation,graves2012supervised}. Moreover, these methods merely identify associations and not causation.

Given the limitations of using input gradients to model the importance of discrete features, we also investigated feature importance using a permutation based sensitivity analysis \citep{fisher2018importance, breiman2001random}. In the test set, we randomly permuted each covariate at each specific time period and measured predictive performance. By permuting each covariate in turn, we destroy any information that a particular covariate provides. If performance then drops significantly relative to a non-permuted baseline, we conclude the feature was important. To prevent correlated variables from leaking information, we simultaneously permuted variables with a correlation coefficient $\ge0.95$. We permuted grouped features within periods of 12 hours to encourage consistency of perturbation along time. \textbf{Figure} \ref{fig:perm_saliency} plots this measure of feature importance over time. Overall, we observe similar trends to the input gradient analysis. In addition to there being greater variability in the first part of the visit, we also observed significant changes in the importance of certain features (measured by sum of importance across time). For example, for the task of predicting in-hospital mortality, respiratory rate is initially the most important feature, but then temperature becomes more important as the patient state evolves. For ARF, a variable pertaining to the Glasgow coma scale is initially most important, before yielding to respiratory rate. 

\vspace{2em}
\begin{figure}[htbp]
\floatconts
{fig:perm_saliency}% label for whole figure
{\caption{Permutation based saliency map of \texttt{mixLSTM-$2$} on three tasks. Each plot shows AUROC degradation for permuting a feature. A larger decrease in AUROC means that the feature is more important with respect to the prediction task. Some noticeable temporal patterns include an increased variability during the first 12 hours which may be a reflection of increased physiological change a patient may experience at the beginning of their ICU stay when interventions are more frequent.}}% caption for whole figure
{%
\subfigure[ARF]{%
\label{fig:perm_saliencyARF}% label for this sub-figure
\includegraphics[width=0.31\linewidth]{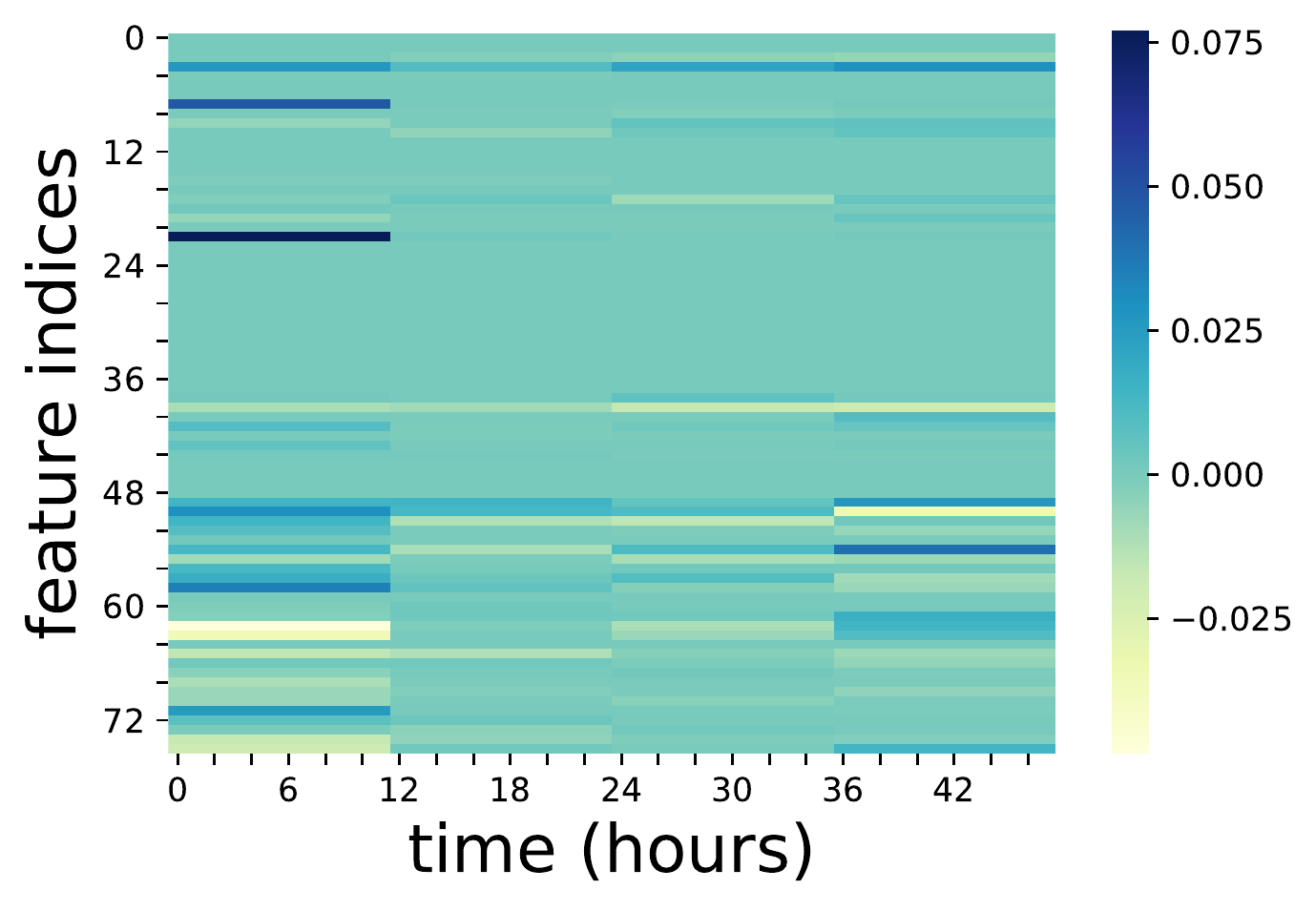}
}%\qquad % space out the images a bit
\subfigure[shock]{%
\label{fig:perm_saliencyShock}% label for this sub-figure
\includegraphics[width=0.31\linewidth]{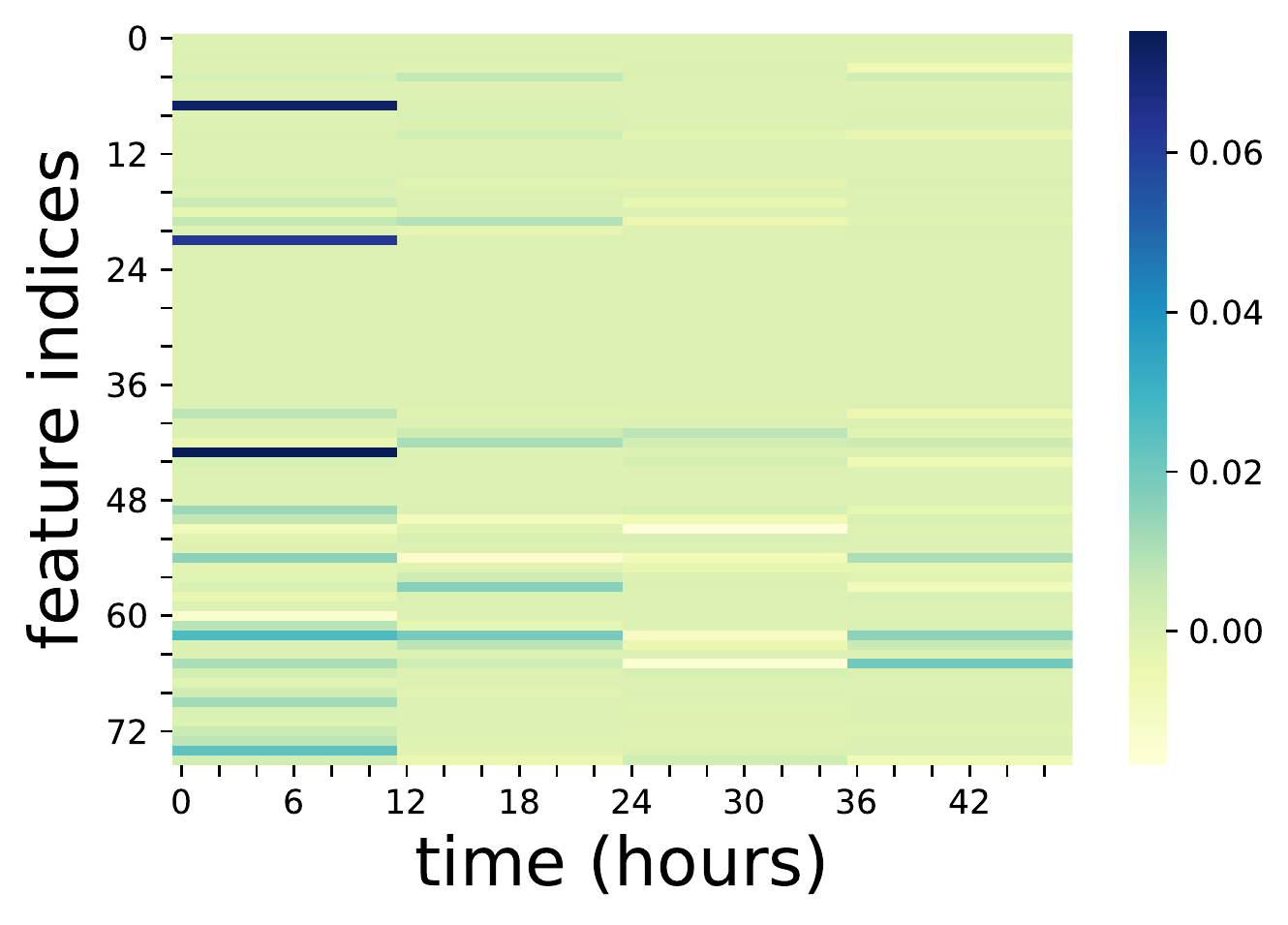}
}%\qquad % space out the images a bit
\subfigure[mortality]{%
\label{fig:perm_saliencymortality}% label for this sub-figure
\includegraphics[width=0.31\linewidth]{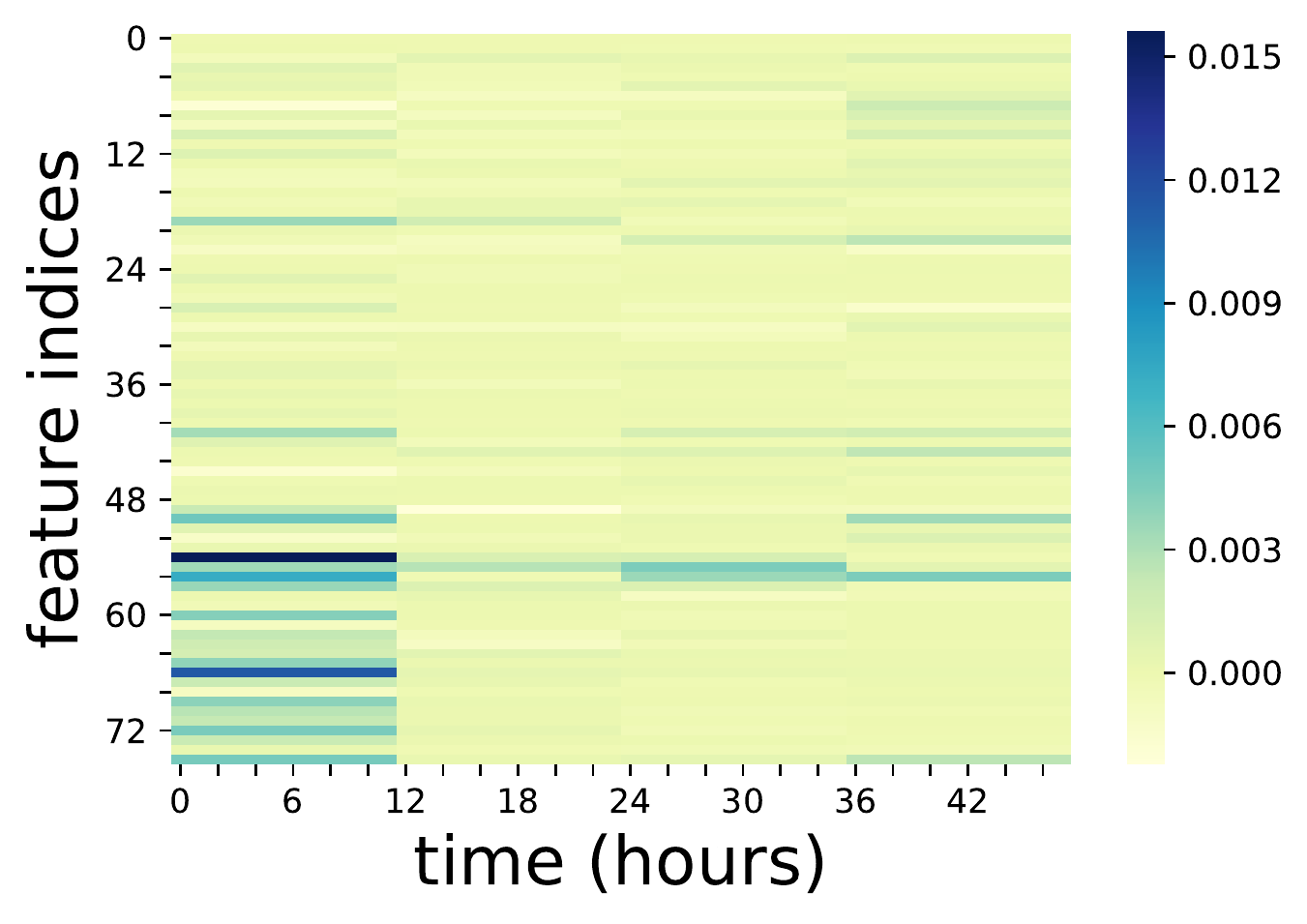}
}
}
\end{figure}

\section{Conclusion}

In this work, we present and explore the issue of temporal conditional shift in clinical time-series data. In addition, we propose a mixture of LSTM model (\texttt{mixLSTM}) and demonstrate that it effectively adapts to scenarios exhibiting temporal conditional shift, consistently outperforming baselines on synthetic and clinical data tasks. We also show that the \texttt{mixLSTM} model can adapt to settings with limited training data and learns meaningful, time-varying relationships from the data.

While \texttt{mixLSTM} achieves consistently better performance on all tasks considered, we note some important limitations. First, we only considered fixed-length datasets. It would be beneficial to compare LSTM and \texttt{mixLSTM}'s ability to generalize to variable length data. Second, our features are largely physiological (e.g., heart rate, temperature). We hypothesize that other types of features such as medications may exhibit stronger time-varying relationships. Third, while it is reasonable to set time zero as the time of ICU admission, patients are admitted to the ICU at different points during the natural history of their illness. Future work should consider the alignment of patient time steps (e.g., learning an individualized model). %Lastly, our current implementation of \texttt{mixLSTM} dynamically creates a new cell at each time step, resulting in an inefficient use of memory.

Despite these limitations, our results suggest that temporal conditional shift is an important aspect of clinical time-series prediction and future work could benefit from considering this problem setting. Our proposed \texttt{mixLSTM} presents a strong starting point from which future work can build. 

%Future work:
%1. individualized mixture model (just learn a attention on blocks for each person; in the memory consuming case, can learn to vote for which one to use in a batch)
%2. architecture search direction (search for shared parameters, instead of architecture)

% ACKNOWLEDGEMENTS ONLY GO IN THE CAMERA-READY, NOT THE SUBMISSION
\pagebreak
\acks{This work was supported by the Michigan Institute for Data Science (MIDAS)\footnote{\url{http://midas.umich.edu/}}, the National Science Foundation (NSF award no. IIS-1553146), and the National Institute of Allergy and Infectious Diseases of the National Institutes of Health (grant no. U01AI124255). The views and conclusions in this document are those of the authors and should not be interpreted as necessarily representing the official policies, either expressed or implied, of the Michigan Institute for Data Science, the National Science Foundation, nor the National Institute of Allergy and Infectious Diseases of the National Institutes of Health.}

%%%%%%%%%%%%%%%
%% Use \bibliography{main} for Overleaf
\bibliography{main}

\begin{thebibliography}{43}
\providecommand{\natexlab}[1]{#1}
\providecommand{\url}[1]{\texttt{#1}}
\expandafter\ifx\csname urlstyle\endcsname\relax
  \providecommand{\doi}[1]{doi: #1}\else
  \providecommand{\doi}{doi: \begingroup \urlstyle{rm}\Url}\fi

\bibitem[Arjovsky et~al.(2016)Arjovsky, Shah, and Bengio]{arjovsky2016unitary}
Martin Arjovsky, Amar Shah, and Yoshua Bengio.
\newblock Unitary evolution recurrent neural networks.
\newblock In \emph{International Conference on Machine Learning}, pages
  1120--1128, 2016.

\bibitem[Avni et~al.(2015)Avni, Lador, Lev, Leibovici, Paul, and
  Grossman]{avni2015vasopressors}
Tomer Avni, Adi Lador, Shaul Lev, Leonard Leibovici, Mical Paul, and Alon
  Grossman.
\newblock Vasopressors for the treatment of septic shock: systematic review and
  meta-analysis.
\newblock \emph{PloS one}, 10\penalty0 (8):\penalty0 e0129305, 2015.

\bibitem[Bellera et~al.(2010)Bellera, MacGrogan, Debled, de~Lara, Brouste, and
  Mathoulin-P{\'e}lissier]{bellera2010variables}
Carine~A Bellera, Ga{\"e}tan MacGrogan, Marc Debled, Christine~Tunon de~Lara,
  V{\'e}ronique Brouste, and Simone Mathoulin-P{\'e}lissier.
\newblock Variables with time-varying effects and the {C}ox model: some
  statistical concepts illustrated with a prognostic factor study in breast
  cancer.
\newblock \emph{BMC medical research methodology}, 10\penalty0 (1):\penalty0
  20, 2010.

\bibitem[Breiman(2001)]{breiman2001random}
Leo Breiman.
\newblock Random forests.
\newblock \emph{Machine learning}, 45\penalty0 (1):\penalty0 5--32, 2001.

\bibitem[Daum{\'e}~III(2007)]{daume2009frustratingly}
Hal Daum{\'e}~III.
\newblock Frustratingly easy domain adaptation.
\newblock \emph{Proc. 45th Ann. Meeting of the Assoc. Computational
  Linguistics}, 2007.

\bibitem[Dekker et~al.(2008)Dekker, De~Mutsert, Van~Dijk, Zoccali, and
  Jager]{dekker2008survival}
Friedo~W Dekker, Ren{\'e}e De~Mutsert, Paul~C Van~Dijk, Carmine Zoccali, and
  Kitty~J Jager.
\newblock Survival analysis: time-dependent effects and time-varying risk
  factors.
\newblock \emph{Kidney international}, 74\penalty0 (8):\penalty0 994--997,
  2008.

\bibitem[Ding et~al.(2017)Ding, Yu, and Jiang]{ding2017recurrent}
Ying Ding, Jianfei Yu, and Jing Jiang.
\newblock Recurrent neural networks with auxiliary labels for cross-domain
  opinion target extraction.
\newblock In \emph{Thirty-First AAAI Conference on Artificial Intelligence},
  2017.

\bibitem[dos Reis et~al.(2016)dos Reis, Flach, Matwin, and
  Batista]{dos2016fast}
Denis~Moreira dos Reis, Peter Flach, Stan Matwin, and Gustavo Batista.
\newblock Fast unsupervised online drift detection using incremental
  kolmogorov-smirnov test.
\newblock In \emph{Proceedings of the 22nd ACM SIGKDD International Conference
  on Knowledge Discovery and Data Mining}, pages 1545--1554. ACM, 2016.

\bibitem[Eigen et~al.(2014)Eigen, Ranzato, and Sutskever]{eigen2013learning}
David Eigen, Marc'Aurelio Ranzato, and Ilya Sutskever.
\newblock Learning factored representations in a deep mixture of experts.
\newblock \emph{International Conference on Learning Representations workshop},
  2014.

\bibitem[Fisher et~al.(2018)Fisher, Rudin, and Dominici]{fisher2018importance}
Aaron Fisher, Cynthia Rudin, and Francesca Dominici.
\newblock All models are wrong but many are useful: Variable importance for
  black-box, proprietary, or misspecified prediction models, using model class
  reliance.
\newblock \emph{arXiv preprint arXiv:1801.01489}, 2018.

\bibitem[Fiterau et~al.(2017)Fiterau, Bhooshan, Fries, Bournhonesque, Hicks,
  Halilaj, Re, and Delp]{fiterau2017shortfuse}
Madalina Fiterau, Suvrat Bhooshan, Jason Fries, Charles Bournhonesque, Jennifer
  Hicks, Eni Halilaj, Christopher Re, and Scott Delp.
\newblock Shortfuse: Biomedical time series representations in the presence of
  structured information.
\newblock In \emph{Machine Learning for Healthcare Conference}, pages 59--74,
  2017.

\bibitem[Gaieski and Mikkelsen(2016)]{gaieski2016shock}
David~F Gaieski and ME~Mikkelsen.
\newblock Definition, classification, etiology, and pathophysiology of shock in
  adults.
\newblock \emph{UpToDate, Waltham, MA. Accesed}, 8:\penalty0 17, 2016.

\bibitem[Ghorbani et~al.(2019)Ghorbani, Abid, and
  Zou]{ghorbani2017interpretation}
Amirata Ghorbani, Abubakar Abid, and James Zou.
\newblock Interpretation of neural networks is fragile.
\newblock \emph{AAAI}, 2019.

\bibitem[Glorot et~al.(2011)Glorot, Bordes, and Bengio]{glorot2011domain}
Xavier Glorot, Antoine Bordes, and Yoshua Bengio.
\newblock Domain adaptation for large-scale sentiment classification: A deep
  learning approach.
\newblock In \emph{Proceedings of the 28th international conference on machine
  learning (ICML-11)}, pages 513--520, 2011.

\bibitem[Gong et~al.(2016)Gong, Zhang, Liu, Tao, Glymour, and
  Sch{\"o}lkopf]{gong2016domain}
Mingming Gong, Kun Zhang, Tongliang Liu, Dacheng Tao, Clark Glymour, and
  Bernhard Sch{\"o}lkopf.
\newblock Domain adaptation with conditional transferable components.
\newblock In \emph{International conference on machine learning}, pages
  2839--2848, 2016.

\bibitem[Graves(2012)]{graves2012supervised}
Alex Graves.
\newblock Supervised sequence labelling.
\newblock In \emph{Supervised sequence labelling with recurrent neural
  networks}, pages 5--13. Springer, 2012.

\bibitem[Ha et~al.(2017)Ha, Dai, and Le]{ha2016hypernetworks}
David Ha, Andrew Dai, and Quoc~V Le.
\newblock Hypernetworks.
\newblock \emph{International Conference on Learning Representations}, 2017.

\bibitem[Harutyunyan et~al.(2019)Harutyunyan, Khachatrian, Kale, Ver~Steeg, and
  Galstyan]{harutyunyan2017multitask}
Hrayr Harutyunyan, Hrant Khachatrian, David~C Kale, Greg Ver~Steeg, and Aram
  Galstyan.
\newblock Multitask learning and benchmarking with clinical time series data.
\newblock \emph{Scientific data}, 6\penalty0 (1):\penalty0 96, 2019.

\bibitem[Johnson et~al.(2016)Johnson, Pollard, Shen, Lehman, Feng, Ghassemi,
  Moody, Szolovits, Celi, and Mark]{johnson2016mimic}
Alistair~EW Johnson, Tom~J Pollard, Lu~Shen, Li-wei~H Lehman, Mengling Feng,
  Mohammad Ghassemi, Benjamin Moody, Peter Szolovits, Leo~Anthony Celi, and
  Roger~G Mark.
\newblock Mimic-iii, a freely accessible critical care database.
\newblock \emph{Scientific data}, 3, 2016.

\bibitem[Kingma and Ba(2015)]{kingma2015adam}
Diederik Kingma and Jimmy Ba.
\newblock Adam: a method for stochastic optimization (2014).
\newblock \emph{International Conference on Learning Representations}, 15,
  2015.

\bibitem[Kohlmorgen et~al.(1998)Kohlmorgen, M{\"u}ller, and
  Pawelzik]{kohlmorgen1998analysis}
Jens Kohlmorgen, Klaus-Robert M{\"u}ller, and Klaus Pawelzik.
\newblock Analysis of drifting dynamics with neural network hidden markov
  models.
\newblock In \emph{Advances in Neural Information Processing Systems}, pages
  735--741, 1998.

\bibitem[Lipton et~al.(2016)Lipton, Kale, Elkan, and
  Wetzel]{lipton2015learning}
Zachary~C Lipton, David~C Kale, Charles Elkan, and Randall Wetzel.
\newblock Learning to diagnose with lstm recurrent neural networks.
\newblock \emph{International Conference on Learning Representations}, 2016.

\bibitem[Ma et~al.(2018)Ma, Zhao, Yi, Chen, Hong, and Chi]{ma2018modeling}
Jiaqi Ma, Zhe Zhao, Xinyang Yi, Jilin Chen, Lichan Hong, and Ed~H Chi.
\newblock Modeling task relationships in multi-task learning with multi-gate
  mixture-of-experts.
\newblock In \emph{Proceedings of the 24th ACM SIGKDD International Conference
  on Knowledge Discovery \& Data Mining}, pages 1930--1939. ACM, 2018.

\bibitem[Meduri et~al.(1996)Meduri, Turner, Abou-Shala, Wunderink, and
  Tolley]{meduri1996noninvasive}
G~Umberto Meduri, Robert~E Turner, Nabil Abou-Shala, Richard Wunderink, and
  Elizabeth Tolley.
\newblock Noninvasive positive pressure ventilation via face mask: first-line
  intervention in patients with acute hypercapnic and hypoxemic respiratory
  failure.
\newblock \emph{Chest}, 109\penalty0 (1):\penalty0 179--193, 1996.

\bibitem[Pan and Yang(2010)]{pan2010survey}
Sinno~Jialin Pan and Qiang Yang.
\newblock A survey on transfer learning.
\newblock \emph{IEEE Transactions on knowledge and data engineering},
  22\penalty0 (10):\penalty0 1345--1359, 2010.

\bibitem[Park and Yoo(2017)]{park2017early}
Hyunsin Park and Chang~D Yoo.
\newblock Early improving recurrent elastic highway network.
\newblock \emph{arXiv preprint arXiv:1708.04116}, 2017.

\bibitem[Rajkomar et~al.(2018)Rajkomar, Oren, Chen, Dai, Hajaj, Hardt, Liu,
  Liu, Marcus, Sun, et~al.]{rajkomar2018scalable}
Alvin Rajkomar, Eyal Oren, Kai Chen, Andrew~M Dai, Nissan Hajaj, Michaela
  Hardt, Peter~J Liu, Xiaobing Liu, Jake Marcus, Mimi Sun, et~al.
\newblock Scalable and accurate deep learning with electronic health records.
\newblock \emph{NPJ Digital Medicine}, 1\penalty0 (1):\penalty0 18, 2018.

\bibitem[Reddi et~al.(2015)Reddi, Poczos, and Smola]{reddi2015doubly}
Sashank~Jakkam Reddi, Barnabas Poczos, and Alex Smola.
\newblock Doubly robust covariate shift correction.
\newblock In \emph{Twenty-Ninth AAAI Conference on Artificial Intelligence},
  2015.

\bibitem[Ross et~al.(2017)Ross, Hughes, and Doshi-Velez]{ross2017right}
Andrew~Slavin Ross, Michael~C Hughes, and Finale Doshi-Velez.
\newblock Right for the right reasons: Training differentiable models by
  constraining their explanations.
\newblock \emph{International Joint Conferences on Artificial Intelligence
  Organization}, 2017.

\bibitem[Savarese and Maire(2019)]{savarese2018learning}
Pedro Savarese and Michael Maire.
\newblock Learning implicitly recurrent {CNN}s through parameter sharing.
\newblock In \emph{International Conference on Learning Representations}, 2019.
\newblock URL \url{https://openreview.net/forum?id=rJgYxn09Fm}.

\bibitem[Selvaraju et~al.(2016)Selvaraju, Cogswell, Das, Vedantam, Parikh,
  Batra, et~al.]{selvaraju2016grad}
Ramprasaath~R Selvaraju, Michael Cogswell, Abhishek Das, Ramakrishna Vedantam,
  Devi Parikh, Dhruv Batra, et~al.
\newblock Grad-cam: Visual explanations from deep networks via gradient-based
  localization., in ‘iccv’, 2016.

\bibitem[Sharif~Razavian et~al.(2014)Sharif~Razavian, Azizpour, Sullivan, and
  Carlsson]{sharif2014cnn}
Ali Sharif~Razavian, Hossein Azizpour, Josephine Sullivan, and Stefan Carlsson.
\newblock Cnn features off-the-shelf: an astounding baseline for recognition.
\newblock In \emph{Proceedings of the IEEE conference on computer vision and
  pattern recognition workshops}, pages 806--813, 2014.

\bibitem[Soemers et~al.(2018)Soemers, Brys, Driessens, Winands, and
  Now{\'e}]{soemers2018adapting}
Dennis~JNJ Soemers, Tim Brys, Kurt Driessens, Mark~HM Winands, and Ann
  Now{\'e}.
\newblock Adapting to concept drift in credit card transaction data streams
  using contextual bandits and decision trees.
\newblock In \emph{Thirty-Second AAAI Conference on Artificial Intelligence},
  2018.

\bibitem[Stefan et~al.(2013)Stefan, Shieh, Pekow, Rothberg, Steingrub, Lagu,
  and Lindenauer]{stefan2013epidemiology}
Mihaela~S Stefan, Meng-Shiou Shieh, Penelope~S Pekow, Michael~B Rothberg, Jay~S
  Steingrub, Tara Lagu, and Peter~K Lindenauer.
\newblock Epidemiology and outcomes of acute respiratory failure in the united
  states, 2001 to 2009: A national survey.
\newblock \emph{Journal of hospital medicine}, 8\penalty0 (2):\penalty0 76--82,
  2013.

\bibitem[Sugiyama et~al.(2007)Sugiyama, Krauledat, and
  M{\~A}{\v{z}}ller]{sugiyama2007covariate}
Masashi Sugiyama, Matthias Krauledat, and Klaus-Robert M{\~A}{\v{z}}ller.
\newblock Covariate shift adaptation by importance weighted cross validation.
\newblock \emph{Journal of Machine Learning Research}, 8\penalty0
  (May):\penalty0 985--1005, 2007.

\bibitem[Tan et~al.(2016)Tan, Qian, and Yu]{tan2016cluster}
Tian Tan, Yanmin Qian, and Kai Yu.
\newblock Cluster adaptive training for deep neural network based acoustic
  model.
\newblock \emph{IEEE/ACM Transactions on Audio, Speech and Language Processing
  (TASLP)}, 24\penalty0 (3):\penalty0 459--468, 2016.

\bibitem[Thiagarajan et~al.(2018)Thiagarajan, Rajan, and
  Sattigeri]{thiagarajan2018can}
Jayaraman~J Thiagarajan, Deepta Rajan, and Prasanna Sattigeri.
\newblock Can deep clinical models handle real-world domain shifts?
\newblock \emph{arXiv preprint arXiv:1809.07806}, 2018.

\bibitem[Van~Hasselt et~al.(2016)Van~Hasselt, Guez, and Silver]{van2016deep}
Hado Van~Hasselt, Arthur Guez, and David Silver.
\newblock Deep reinforcement learning with double q-learning.
\newblock In \emph{Thirtieth AAAI Conference on Artificial Intelligence}, 2016.

\bibitem[Vaswani et~al.(2017)Vaswani, Shazeer, Parmar, Uszkoreit, Jones, Gomez,
  Kaiser, and Polosukhin]{vaswani2017attention}
Ashish Vaswani, Noam Shazeer, Niki Parmar, Jakob Uszkoreit, Llion Jones,
  Aidan~N Gomez, {\L}ukasz Kaiser, and Illia Polosukhin.
\newblock Attention is all you need.
\newblock In \emph{Advances in Neural Information Processing Systems}, pages
  5998--6008, 2017.

\bibitem[Wang et~al.(2018)Wang, Yu, Dou, Darrell, and
  Gonzalez]{wang2018skipnet}
Xin Wang, Fisher Yu, Zi-Yi Dou, Trevor Darrell, and Joseph~E Gonzalez.
\newblock Skipnet: Learning dynamic routing in convolutional networks.
\newblock In \emph{Proceedings of the European Conference on Computer Vision
  (ECCV)}, pages 409--424, 2018.

\bibitem[Wiens et~al.(2016)Wiens, Guttag, and Horvitz]{wiens2016patient}
Jenna Wiens, John Guttag, and Eric Horvitz.
\newblock Patient risk stratification with time-varying parameters: a multitask
  learning approach.
\newblock \emph{The Journal of Machine Learning Research}, 17\penalty0
  (1):\penalty0 2797--2819, 2016.

\bibitem[Zhang et~al.(2013)Zhang, Sch{\"o}lkopf, Muandet, and
  Wang]{zhang2013domain}
Kun Zhang, Bernhard Sch{\"o}lkopf, Krikamol Muandet, and Zhikun Wang.
\newblock Domain adaptation under target and conditional shift.
\newblock In \emph{International Conference on Machine Learning}, pages
  819--827, 2013.

\bibitem[Zhuang et~al.(2015)Zhuang, Cheng, Luo, Pan, and
  He]{zhuang2015supervised}
Fuzhen Zhuang, Xiaohu Cheng, Ping Luo, Sinno~Jialin Pan, and Qing He.
\newblock Supervised representation learning: Transfer learning with deep
  autoencoders.
\newblock In \emph{Twenty-Fourth International Joint Conference on Artificial
  Intelligence}, 2015.

\end{thebibliography}
%% Use \input{main.bbl} for uploading to arxiv (also need to manually update the bbl file)
% \input{main.bbl}
%%%%%%%%%%%%%%%

\appendix
\section{Details of Data \& Features}\label{appendix:a}

% Some more details about those methods, so we can actually reproduce them.  After the blind review period, you could link to a repository for the code also.  

\begin{table}[H]\scriptsize
\caption{The 17 physiological features extracted from MIMIC-III database, the source tables, and the corresponding ITEMIDs}\label{tab:17vars}
\hskip-1.5cm\begin{tabular}{llll}
\toprule
Index & Variable Name                      & Table(s)              & ITEMID(s)                                                                                \\
\midrule
1     & Capillary refill rate              & CHARTEVENTS           & 3348, 115, 8377                                                                          \\
2     & Diastolic blood pressure           & CHARTEVENTS           & \makecell[l]{8368, 220051, 225310, 8555, 8441, 220180, 8502, \\ 8440, 8503, 8504, 8507, 8506, 224643}     \\
3     & Fraction inspired oxygen           & CHARTEVENTS           & 3420, 223835, 3422, 189, 727                                                             \\
4     & Glascow coma scale eye opening     & CHARTEVENTS           & 184, 220739                                                                              \\
5     & Glascow coma scale motor response  & CHARTEVENTS           & 454, 223901                                                                              \\
6     & Glascow coma scale total           & CHARTEVENTS           & 198,                                                                                     \\
7     & Glascow coma scale verbal response & CHARTEVENTS           & 723, 223900                                                                              \\
8     & Glucose                            & CHARTEVENTS/LABEVENTS & \makecell[l]{50931, 807, 811, 1529, 50809, 51478, 3745, 225664, \\220621, 226537}                        \\
9     & Heart Rate                         & CHARTEVENTS           & 221, 220045                                                                              \\
10    & Height                             & CHARTEVENTS           & 226707, 226730, 1394                                                                     \\
11    & Mean blood pressure                & CHARTEVENTS           & \makecell[l]{52, 220052, 225312, 224, 6702, 224322, 456, \\220181, 3312, 3314, 3316, 3322, 3320}         \\
12    & Oxygen saturation                  & CHARTEVENTS/LABEVENTS & 834, 50817, 8498, 220227, 646, 220277                                                    \\
13    & Respiratory rate                   & CHARTEVENTS           & 618, 220210, 3603, 224689, 614, 651, 224422, 615, 224690                                 \\
14    & Systolic blood pressure            & CHARTEVENTS           & \makecell[l]{51, 220050, 225309, 6701, 455, 220179, 3313, \\3315, 442, 3317, 3323, 3321, 224167, 227243} \\
15    & Temperature                        & CHARTEVENTS           & 3655, 677, 676, 223762, 3654, 678, 223761, 679                                           \\
16    & Weight                             & CHARTEVENTS           & 763, 224639, 226512, 3580, 3693, 3581, 226531, 3582                                      \\
17    & pH                                 & CHARTEVENTS/LABEVENTS & \makecell[l]{50820, 51491, 3839, 1673, 50831, 51094, 780, \\1126, 223830, 4753, 4202, 860, 220274}       \\
\bottomrule
\end{tabular}
\end{table}

{\footnotesize
\begin{longtable}{lll}
\caption{The 76 time-series features used as input to all the models. }\label{tab:76features}\\
\toprule
Index & Feature Name                                                             & Type    \\
\midrule
0     & Capillary refill rate-\textgreater{}0.0                                  & Binary  \\
1     & Capillary refill rate-\textgreater{}1.0                                  & Binary  \\
2     & Diastolic blood pressure                                                 & Numeric \\
3     & Fraction inspired oxygen                                                 & Numeric \\
4     & Glascow coma scale eye opening-\textgreater{}To Pain                     & Binary  \\
5     & Glascow coma scale eye opening-\textgreater{}3 To speech                 & Binary  \\
6     & Glascow coma scale eye opening-\textgreater{}1 No Response               & Binary  \\
7     & Glascow coma scale eye opening-\textgreater{}4 Spontaneously             & Binary  \\
8     & Glascow coma scale eye opening-\textgreater{}None                        & Binary  \\
9     & Glascow coma scale eye opening-\textgreater{}To Speech                   & Binary  \\
10    & Glascow coma scale eye opening-\textgreater{}Spontaneously               & Binary  \\
11    & Glascow coma scale eye opening-\textgreater{}2 To pain                   & Binary  \\
12    & Glascow coma scale motor response-\textgreater{}1 No Response            & Binary  \\
13    & Glascow coma scale motor response-\textgreater{}3 Abnorm flexion         & Binary  \\
14    & Glascow coma scale motor response-\textgreater{}Abnormal extension       & Binary  \\
15    & Glascow coma scale motor response-\textgreater{}No response              & Binary  \\
16    & Glascow coma scale motor response-\textgreater{}4 Flex-withdraws         & Binary  \\
17    & Glascow coma scale motor response-\textgreater{}Localizes Pain           & Binary  \\
18    & Glascow coma scale motor response-\textgreater{}Flex-withdraws           & Binary  \\
19    & Glascow coma scale motor response-\textgreater{}Obeys Commands           & Binary  \\
20    & Glascow coma scale motor response-\textgreater{}Abnormal Flexion         & Binary  \\
21    & Glascow coma scale motor response-\textgreater{}6 Obeys Commands         & Binary  \\
22    & Glascow coma scale motor response-\textgreater{}5 Localizes Pain         & Binary  \\
23    & Glascow coma scale motor response-\textgreater{}2 Abnorm extensn         & Binary  \\
24    & Glascow coma scale total-\textgreater{}11                                & Binary  \\
25    & Glascow coma scale total-\textgreater{}10                                & Binary  \\
26    & Glascow coma scale total-\textgreater{}13                                & Binary  \\
27    & Glascow coma scale total-\textgreater{}12                                & Binary  \\
28    & Glascow coma scale total-\textgreater{}15                                & Binary  \\
29    & Glascow coma scale total-\textgreater{}14                                & Binary  \\
30    & Glascow coma scale total-\textgreater{}3                                 & Binary  \\
31    & Glascow coma scale total-\textgreater{}5                                 & Binary  \\
32    & Glascow coma scale total-\textgreater{}4                                 & Binary  \\
33    & Glascow coma scale total-\textgreater{}7                                 & Binary  \\
34    & Glascow coma scale total-\textgreater{}6                                 & Binary  \\
35    & Glascow coma scale total-\textgreater{}9                                 & Binary  \\
36    & Glascow coma scale total-\textgreater{}8                                 & Binary  \\
37    & Glascow coma scale verbal response-\textgreater{}1 No Response           & Binary  \\
38    & Glascow coma scale verbal response-\textgreater{}No Response             & Binary  \\
39    & Glascow coma scale verbal response-\textgreater{}Confused                & Binary  \\
40    & Glascow coma scale verbal response-\textgreater{}Inappropriate Words     & Binary  \\
41    & Glascow coma scale verbal response-\textgreater{}Oriented                & Binary  \\
42    & Glascow coma scale verbal response-\textgreater{}No Response-ETT         & Binary  \\
43    & Glascow coma scale verbal response-\textgreater{}5 Oriented              & Binary  \\
44    & Glascow coma scale verbal response-\textgreater{}Incomprehensible sounds & Binary  \\
45    & Glascow coma scale verbal response-\textgreater{}1.0 ET/Trach            & Binary  \\
46    & Glascow coma scale verbal response-\textgreater{}4 Confused              & Binary  \\
47    & Glascow coma scale verbal response-\textgreater{}2 Incomp sounds         & Binary  \\
48    & Glascow coma scale verbal response-\textgreater{}3 Inapprop words        & Binary  \\
49    & Glucose                                                                  & Numeric \\
50    & Heart Rate                                                               & Numeric \\
51    & Height                                                                   & Numeric \\
52    & Mean blood pressure                                                      & Numeric \\
53    & Oxygen saturation                                                        & Numeric \\
54    & Respiratory rate                                                         & Numeric \\
55    & Systolic blood pressure                                                  & Numeric \\
56    & Temperature                                                              & Numeric \\
57    & Weight                                                                   & Numeric \\
58    & pH                                                                       & Numeric \\
59    & mask-\textgreater{}Capillary refill rate                                 & Binary  \\
60    & mask-\textgreater{}Diastolic blood pressure                              & Binary  \\
61    & mask-\textgreater{}Fraction inspired oxygen                              & Binary  \\
62    & mask-\textgreater{}Glascow coma scale eye opening                        & Binary  \\
63    & mask-\textgreater{}Glascow coma scale motor response                     & Binary  \\
64    & mask-\textgreater{}Glascow coma scale total                              & Binary  \\
65    & mask-\textgreater{}Glascow coma scale verbal response                    & Binary  \\
66    & mask-\textgreater{}Glucose                                               & Binary  \\
67    & mask-\textgreater{}Heart Rate                                            & Binary  \\
68    & mask-\textgreater{}Height                                                & Binary  \\
69    & mask-\textgreater{}Mean blood pressure                                   & Binary  \\
70    & mask-\textgreater{}Oxygen saturation                                     & Binary  \\
71    & mask-\textgreater{}Respiratory rate                                      & Binary  \\
72    & mask-\textgreater{}Systolic blood pressure                               & Binary  \\
73    & mask-\textgreater{}Temperature                                           & Binary  \\
74    & mask-\textgreater{}Weight                                                & Binary  \\
75    & mask-\textgreater{}pH                                                    & Binary  \\
\bottomrule
\end{longtable}
}
\vspace{-2em}

\end{document}